\documentclass[10pt,twocolumn,letterpaper]{article}

\usepackage{iccv}
\usepackage{times}
\usepackage{epsfig}
\usepackage{graphicx}
\usepackage{amsmath}
\usepackage{amssymb}

\usepackage[pagebackref=true,breaklinks=true,letterpaper=true,colorlinks,bookmarks=false]{hyperref}

\iccvfinalcopy %

\usepackage[accsupp]{axessibility}
\usepackage{color}
\usepackage{booktabs}
\usepackage{multirow}
\usepackage[official]{eurosym}
\usepackage{mathrsfs}
\usepackage{yfonts}
\usepackage{xspace}
\usepackage{pifont} %
\usepackage{enumitem}
\usepackage[numbers,sort]{natbib}
\usepackage{epigraph} 
\usepackage{diagbox} %

\newcommand{\methodshort}{\!\,\texttt{TMR\xspace}}

\newcommand{\cam}[1]{\textcolor{black}{#1}}

\definecolor{aliceblue}{rgb}{0.94, 0.97, 1.0}

\newcommand{\cmark}{\ding{51}}%
\newcommand{\xmark}{\ding{55}}%
\usepackage[subtle]{savetrees} %
\def\sepappendix{0}

\def\withaction{0}

\begin{document}

\title{\methodshort{}: Text-to-Motion Retrieval Using Contrastive 3D Human Motion Synthesis}

\author{Mathis Petrovich$^{1,2}$ \quad Michael J. Black$^{2}$ \quad G\"ul Varol$^1$\\
$^{1}$ LIGM, \'Ecole des Ponts, Univ Gustave Eiffel, CNRS, France \\
$^{2}$ Max Planck Institute for Intelligent Systems, T\"{u}bingen, Germany \\
{\tt\small \url{https://mathis.petrovich.fr/tmr} }
}

\maketitle

\begin{abstract}
In this paper, we present \methodshort{}, a simple yet effective approach for text to 3D human motion retrieval. While previous work has only treated retrieval as a proxy evaluation metric, we tackle it as a standalone task.
Our method extends the state-of-the-art text-to-motion synthesis model TEMOS, and incorporates a contrastive loss to better structure the cross-modal latent space. We show that maintaining the motion generation loss, along with the contrastive training, is crucial to obtain good performance. We introduce a benchmark for evaluation and provide an in-depth analysis by reporting results on several protocols. Our extensive experiments on the KIT-ML and HumanML3D datasets show that \methodshort{} outperforms the prior work by a significant margin, for example reducing the median rank from 54 to 19. Finally, we showcase the potential of our approach on moment retrieval. Our code and models are publicly available
at \href{https://mathis.petrovich.fr/tmr}{https://mathis.petrovich.fr/tmr}.
\end{abstract}
\section{Introduction}
\label{sec:intro}

\setlength{\epigraphwidth}{0.44\textwidth}
{\scriptsize
\epigraph{\textit{The language of movement cannot be translated into words.}\\
	\hspace{155pt}Barbara Mettler}
{}
}

\begin{figure}
	\centering
	\includegraphics[width=0.99\linewidth]{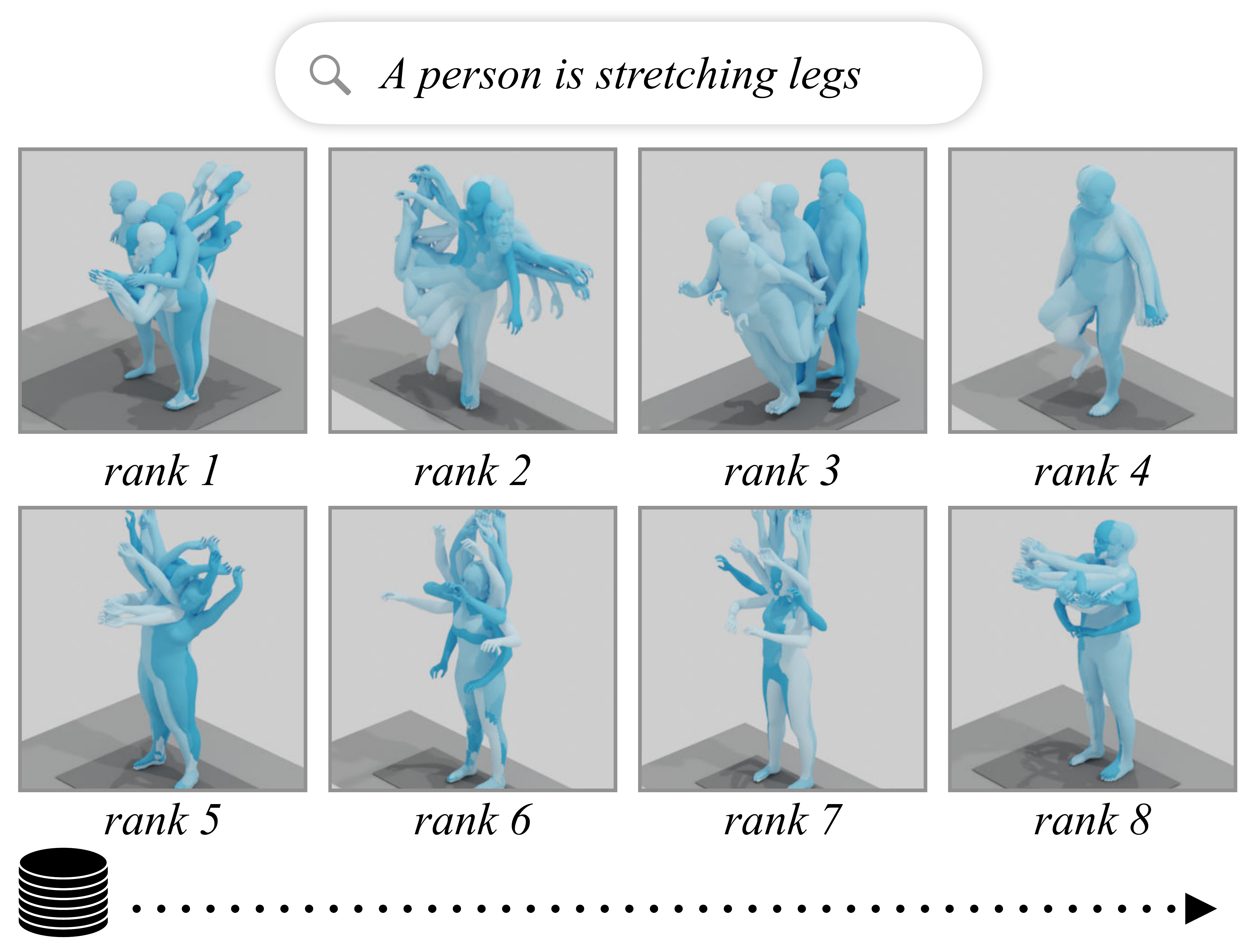}
	\caption{\textbf{Text-to-motion retrieval:}
        We illustrate the task of text-based motion retrieval
        where the goal is to rank a gallery of motions
        according to their similarity to the given query
        in the form of a natural language description.
     }
    \vspace{-0.3cm}
	\label{fig:teaser}
\end{figure}

We ask the question whether a cross-modal space exists between 3D human motions
and language.
Our goal is to retrieve the most relevant 3D human motion from a gallery,
given a natural language query that describes the desired motion (as illustrated
in Figure~\ref{fig:teaser}). 
While text-to-image retrieval is a well-established problem
within the broader vision~\&~language field \cite{clip-pmlr-v139-radford21a},
there has been less focus on the related task of \textit{text-to-motion} retrieval. 
Searching an existing motion capture dataset based on text input can often serve
as a viable alternative to text-to-human-motion synthesis in many applications,
while also providing the added benefit that the retrieved motion is  guaranteed to be realistic.
Additionally,
once a cross-modal embedding is built to map text and motions into a joint representation space,
both of the symmetrical text-to-motion and \textit{motion-to-text}  tasks can be performed.
Such a retrieval-based solution has a range of applications, including
automatically indexing large motion capture collections,
and helping to initialize the cumbersome text labeling process,
by assigning the nearest text to each motion.

Let us first differentiate text-to-motion \textit{retrieval} from text-to-motion \textit{synthesis}.
Motion synthesis \cite{petrovich22temos,athanasiou22teach,tevet2023human,chen2023} involves
generating \textit{new} data samples
that go beyond the existing training set, while motion retrieval searches through
existing motion capture collections.
For certain applications, reusing motions from a collection
may be sufficient, provided the collection is large enough to
contain what the user is searching for. Unlike generative models for motion synthesis, 
which struggle to produce physically plausible, realistic sequences
\cite{petrovich21actor,petrovich22temos,athanasiou22teach},
a retrieval model has the advantage that it always returns a realistic motion.
With this motivation, we pose the problem as a nearest neighbor search
through a cross-modal text-motion space.

Early work performs search through motion databases
to build motion graphs \cite{Kovar2002MotionGraphs,arikan2003siggraph}
by finding paths between existing motions and synthesizing new motions
by stitching motions together with generated transitions.
If the motion database is labeled with actions, the user can specify
a series of actions to combine \cite{arikan2003siggraph}.
In contrast, our search database is \textit{not} labeled with text.
\textit{Motion matching} \cite{buttner2015},
on the other hand, seeks to find the animation that best fits
the current motion by searching a database of animations,
doing motion-to-motion retrieval \cite{Sidenbladh:ECCV:02}.
Our framework fundamentally differs from these lines of work in that
our task is multi-modal, i.e., user query is text,
which is compared against motions.
The most similar work to ours is the very recent
model from Guo~et~al.~\cite{Guo_2022_CVPR},
which trains for a joint embedding space between the two
modalities. This model is only used to provide
a performance measure for motion synthesis tasks,
by querying a generated motion within a gallery of
32 descriptions (i.e., motion-to-text retrieval),
and counting how many times the correct text is retrieved\footnote{
While the paper \cite{Guo_2022_CVPR} describes a motion-to-text retrieval
metric, we notice that the provided code performs text-to-motion retrieval.
}.
While this can be considered as the first text-motion retrieval model
in the literature, its main limitation is the low performance,
in particular when the gallery contains fine-grained descriptions.
We substantially improve over \cite{Guo_2022_CVPR},
by incorporating a joint synthesis \& retrieval framework,
as well as a more powerful contrastive training \cite{oord2018representation}.

We get inspiration from image-text models such as BLIP~\cite{li2022blip} and CoCa~\cite{yu2022coca},
which formulate a multi-task objective. Besides the standard dual-encoder
matching (such as CLIP~\cite{clip-pmlr-v139-radford21a} with two unimodal encoders for image and text),
\cite{li2022blip,yu2022coca} also employ a text synthesis branch, performing image
captioning. Such a generative capability potentially helps the model go beyond
`bag-of-words' understanding of vision-language concepts,
observed for the naive contrastive models \cite{doven2023,yuksekgonul2022}.
In our case, we depart from TEMOS~\cite{petrovich22temos}
which already has a synthesis branch to generate motions from text. We incorporate
a cross-modal contrastive loss (i.e., InfoNCE~\cite{oord2018representation})
in this framework to jointly train text-to-motion synthesis
and text-to-motion retrieval tasks. We empirically demonstrate
significant improvements with this approach
when ablating the importance of each task.

Text-motion data differs from its text-image counterparts
particularly due to the %
nature of motion descriptions.
In fact, for an off-the-shelf large language model, %
sentences describing different motions tend to be similar,
since they fall within the same topic of human motions.
For example,
the text-text cosine similarities \cite{mpnet2020} after encoding motion descriptions from the KIT training set \cite{Plappert2016_KIT_ML}
are on average 0.71 %
on a scale between [0, 1],
while this value is 0.56 (almost orthogonal) %
on a random subset of LAION~\cite{schuhmann2021laion} image descriptions with the same size.
This poses several challenges. Typical motion datasets 
\cite{Plappert2016_KIT_ML,Guo_2022_CVPR,babel2021}
contain similar motions with different accompanying texts, e.g., `person walks', `human walks',
as well as similar texts with different motions, e.g., `walk backwards', `walk forwards'.
With naive contrastive training \cite{oord2018representation}, one would make all samples within a batch
as negatives, except the corresponding label for a given anchor.
In this work, we take into account the fact that there are potentially
significant similarities between pairs within a batch. 
Specifically,
we discard pairs that have a text-text similarity in their labels
that is above a certain threshold. Such careful negative sampling leads to
performance improvements.

\if\withaction1{
In this paper, we illustrate additional use cases for our retrieval model,
and highlight these tasks as potential future avenues for research.
We introduce zero-shot 3D action recognition, where the motion-to-text model
can be applied to a new domain with action labels as the text gallery.
We test the limits of our approach and perform
60-way
action classification on the BABEL dataset \cite{babel2021},
by querying motions
through our model
trained on HumanML3D \cite{Guo_2022_CVPR}.
Despite %
action labels %
having a domain gap with the training set,
we obtain reasonable performance.
A second use case is moment retrieval on long 3D motion sequences.
} \else{
In this paper, we illustrate an additional use case
for our retrieval model -- zero-shot temporal localization --
and highlight this task
as a potential future avenue for research.
} \fi
Similar to temporal localization in videos with natural language
queries~\cite{escorcia2019, tvr2020, hendricks2017, gao2017, regneri2013},
also referred to as ``moment retrieval", we showcase the grounding capability of our model
by directly applying it on long motion sequences to retrieve corresponding moments.
We illustrate results on the BABEL dataset \cite{babel2021}, which typically
contains a series of text annotations for each motion sequence.
\if\withaction0{
Note that the task is zero-shot, because the model has not been
trained for localization, and at the same time has not seen BABEL
labels, which come from a different domain (e.g., typically
action-like descriptions instead of full sentences).
} \fi

Our contributions are the following:
(i) We address the little-studied problem of text-to-motion retrieval, and
introduce a series of evaluation benchmarks with varying difficulty. %
(ii) We propose a joint synthesis and retrieval framework, as well as
negative filtering, and obtain state-of-the-art performance on text-motion retrieval.
(iii) We provide extensive experiments to analyze the effects of each component in controlled settings.
Our code and models are publicly available\footnote{\url{https://mathis.petrovich.fr/tmr}}.

\section{Related work}
\label{sec:relatedwork}

We present an overview of the most closely related work
on text-to-motion synthesis and retrieval,
as well as a brief discussion on cross-modal
retrieval works.

\paragraph{Text and human motion.}
The research on human motion modeling has recently witnessed
an increasing interest in bridging the gap between semantics and
3D human body motions, in particular for text-conditioned motion synthesis.
Unlike unconstrained 3D human motion synthesis
\cite{Yan_2019_ICCV,Zhao2020BayesianAH,Zhang2020PerpetualMG},
action-conditioned \cite{chuan2020action2motion,petrovich21actor},
or text-conditioned \cite{Ghosh_2021_ICCV,petrovich22temos,athanasiou22teach,tevet2023human,chen2023,Zhang2023t2mgpt,Kalakonda2022,Lee2022multiact} 
models add semantic controls to the generation process.
The goal of these works is to generate either deterministically 
\cite{lin2018,Ahuja2019Language2PoseNL,Ghosh_2021_ICCV}
or probabilistically \cite{petrovich22temos}, motion sequences that are faithful to the textual description inputs. 
Note that this is different than gesture synthesis from speech 
\cite{gesturematching22}, in that the text describes the motion content.
However, despite remarkable progress in text-to-image synthesis 
\cite{ramesh21a,rombach2022high},
text-to-motion synthesis remains at a very nascent stage. %
The realism of the synthesized motions is limited, e.g., foot sliding artifacts
\cite{petrovich22temos}. We turn to text-to-motion retrieval as another alternative,
and perhaps complementary approach to obtain motions for a given textual description.
Our focus is therefore
different from the work on synthesis.
However,
we do make use of a motion synthesis branch to aid the retrieval task. 

Motion retrieval is relatively less explored.
As briefly mentioned in Section~\ref{sec:intro},
motion-to-motion retrieval (e.g., motion matching \cite{Sidenbladh:ECCV:02, buttner2015, Holden:2020})
methods exist.
However, the \textit{text}-to-motion
retrieval task is more challenging due to being cross-modal,
i.e., nearest neighbor search across text and motion modalities.
Within 
this category, the very recent work of
Guo~et~al.~\cite{Guo_2022_CVPR} trains a retrieval model purely
for evaluation purposes, and applies a margin-based contrastive loss \cite{hadsell2006},
using Euclidean distance between all pairs within a batch.

Two works are particularly relevant to ours. Firstly, we build on
the TEMOS~\cite{petrovich22temos} text-to-motion synthesis
model, which also has a cross-modal embedding space. However,
text and motion embeddings are encouraged to be similar only
across positive pairs. We therefore add a contrastive training strategy
to incorporate negatives, consequently improving its retrieval capability
from a large gallery of fine-grained motions.
Secondly, we compare to the aforementioned method of \cite{Guo_2022_CVPR},
whose motion-to-text retrieval model is adopted for measuring text-to-motion
generation performance automatically by other works \cite{tevet2023human,zhang2022motiondiffuse,Dabral2022}.

\paragraph{Cross-modal retrieval.}
Among widely adopted vision \& language retrieval models, some successful examples
include
CLIP \cite{clip-pmlr-v139-radford21a}, BLIP~\cite{li2022blip}, and CoCa~\cite{yu2022coca} for images,
and
MIL-NCE \cite{miech20endtoend}, Frozen \cite{Bain_2021_ICCV}, and CLIP4Clip \cite{clip4clip2021} for videos.
They all use variants of cross-modal contrastive learning techniques,
such as InfoNCE~\cite{oord2018representation}, which we also employ in this work.
As discussed in Section~\ref{sec:intro}, we draw inspiration from BLIP~\cite{li2022blip} and CoCa~\cite{yu2022coca},
which add synthesis branches to standard retrieval frameworks.
Our approach is similar in spirit to these works in that we perform a cross-modal vision \& language
retrieval task, but differ in focusing on 3D human motion retrieval,
which, to the best of our knowledge, has not been benchmarked.

\begin{figure*}
	\centering
	\includegraphics[width=0.95\linewidth]{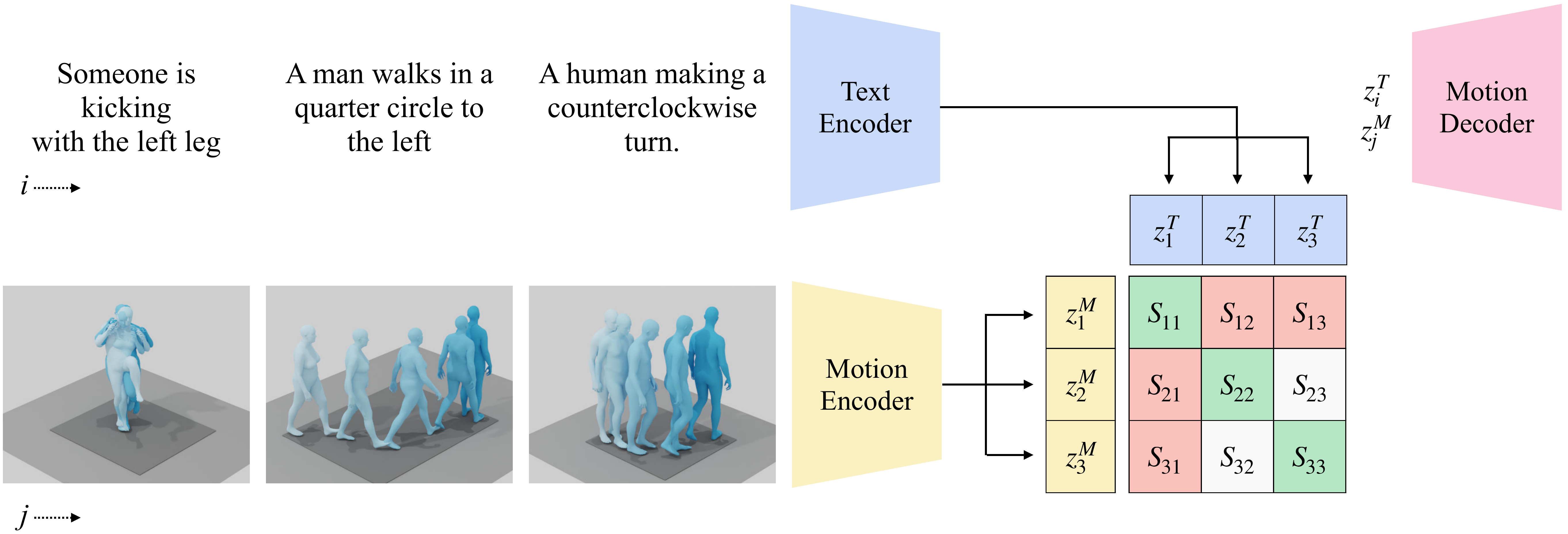}
	\caption{\textbf{Joint motion retrieval and synthesis:} 
        A simplified view of our \methodshort{} framework is
        presented, where we focus on the similarity matrix
        defined between text-motion pairs within a batch.
        Here, we show a batch of 3 samples for illustration purposes.
        The goal of the contrastive objective is to maximize
        the diagonal denoting positive pairs (green),
        and to minimize the off-diagonal negative pair items
        that have text similarity below a threshold (red).
        In this example, remaining similarities
        $S_{23}$ and $S_{32}$ are discarded from the loss
        computation because there is high text similarity between
        $T_2$ and $T_3$.
        The rest of the model remains similar to TEMOS~\cite{petrovich22temos},
        which decodes a motion from both text $z^{T_i}$
        and motion $z^{M_j}$ latent vectors. See text for further details.
	}
	\label{fig:method}
        \vspace{-0.3cm}
\end{figure*}

\section{Text-to-motion retrieval}
\label{sec:method}

In this section, we introduce the task and the terminology associated with text-to-motion retrieval (Section~\ref{subsec:definition}). Next, we present our model, named \methodshort{}, and its training protocol (Section~\ref{subsec:model}). We then explain our simple approach for identifying and filtering incorrect negatives (Section~\ref{subsec:filter}), followed by a discussion of the implementation details (Section~\ref{subsec:implem_details}).

\subsection{Definitions}
\label{subsec:definition}

Given a natural language query $T$, such as 
`\textit{A person walks and then makes a right turn.'}, 
the goal is to rank the motions from a database (i.e., the gallery)
according to their semantic correspondence with the text query,
and to retrieve the motion that matches best to the textual description.
In other words, the task involves sorting the database 
so that the top ranked motions are the most relevant matches, 
i.e., creating a search engine to index motions.
Additionally, we define the symmetric (and complementary) task, 
namely motion-to-text retrieval, where the aim is to retrieve 
the most suitable %
caption 
that matches a 
given 
motion from a database of texts.

\noindent\textbf{3D human motion} refers to a sequence of human poses. 
The task does not impose any limitations on the type of representation used, 
such as joint positions, rotations, or parametric models such as SMPL~\cite{smpl2015}. 
As detailed in Section~\ref{subsec:datasets}, we choose to use the 
representation employed by Guo et al.~\cite{Guo_2022_CVPR} 
to facilitate comparisons with previous work.
\cam{While our experiments are based on the SMPL skeleton, our method is applicable to any skeleton topology.
One could alternatively use retargetting~\cite{Villegas_2018_CVPR,wang2020neural,aberman2020skeleton} to adapt
to other body representations.}

\noindent\textbf{Text description} refers to a sequence of words
describing the action performed by a human in natural language.
We do not restrict the format of the motion description.
The text can be simply an action name (e.g., `walk') or a full sentence
(e.g., `a human is walking').
The sentences can be fine-grained (e.g., `a human is walking in a circle slowly'),
and may contain one or several actions, simultaneously (e.g., `walking while waving') 
or sequentially (e.g., `walking then sitting').

\subsection{Joint training of retrieval and synthesis}
\label{subsec:model}

We introduce \methodshort{},
which extends the Transformer-based text-to-motion synthesis model
TEMOS~\cite{petrovich22temos}
by incorporating additional losses to make it suitable for the retrieval task.
The architecture consists of two independent encoders for
inputting motion and text, as well as a decoder that outputs motion
(see Figure~\ref{fig:method} for an overview).
In the following, we review TEMOS~\cite{petrovich22temos}
components and our added contrastive training.

\noindent\textbf{Dual encoders.} One approach to solving cross-modal retrieval tasks 
involves defining a similarity function
between the two modalities. 
In our case, the two modalities are text and motion.
The similarity function can be applied to compare a given query 
with each element in the database, and the maximum value would indicate the best match.
In this paper, we follow the approach taken by previous work on metric learning, 
such as CLIP~\cite{clip-pmlr-v139-radford21a}, 
by defining one encoder for each modality and then 
computing the cosine similarity between their respective embeddings.
Such dual embedding
has the advantage of fast inference time since the gallery embeddings can be
computed and stored beforehand \cite{miech2021}.

Our model is built upon the components of TEMOS~\cite{petrovich22temos}, 
which already provides a motion encoder and a text encoder, mapping
them to a joint space (building on the idea from Language2Pose~\cite{Ahuja2019Language2PoseNL}); this
serves as a 
strong baseline
for our work. 
Both motion and text encoders are Transformer encoders~\cite{vaswani2017attention} 
with additional learnable distribution parameters, as in the VAE-based ACTOR~\cite{petrovich21actor}. 
They are probabilistic in nature, 
outputting parameters of a Gaussian distribution ($\mu$ and $\Sigma$) 
from which a latent vector $z \in \mathbb{R}^{d}$ can be sampled. 
While the text encoder takes text features 
from a pre-trained and frozen DistilBERT~\cite{distilbert_sanh} network as input, 
the motion sequence is fed directly in the motion encoder.
Note that when performing retrieval, we directly use the output embedding
that corresponds to the mean token ($\mu^M$ for motion, $\mu^T$ for text).

\noindent\textbf{Motion decoder.} TEMOS is trained for the task of motion synthesis 
and comes equipped with a motion decoder branch. 
This decoder is identical to the one used in ACTOR~\cite{petrovich21actor},
which supports variable-duration generation.
More specifically, it takes a latent vector $z \in \mathbb{R}^{d}$
and a sinusoidal positional encoding as input, 
and generates a motion non-autoregressively 
through a single forward pass.
We show in Section~\ref{subsec:sota} that keeping this branch helps improve the results.

\noindent\textbf{TEMOS losses.} We keep the same base set of losses of \cite{petrovich22temos},
defined as the weighted sum 
$\mathcal{L}_{\text{TEMOS}} = \mathcal{L}_{\text{R}} + \lambda_{\text{KL}}\mathcal{L}_{\text{KL}} + \lambda_{\text{E}}\mathcal{L}_{\text{E}}$.
In summary,
a reconstruction loss term $\mathcal{L}_{\text{R}}$ measures the motion reconstruction 
given text or motion input (via a smooth L1 loss).
A Kullback-Leibler (KL) divergence loss term $\mathcal{L}_{\text{KL}}$ is
composed of four losses: two of them to regularize each 
encoded distributions -- $\mathcal{N}(\mu^M, \Sigma^M)$ for motion
and $\mathcal{N}(\mu^T, \Sigma^T)$ for text --
to come from a normal distribution $\mathcal{N}(0, I)$. %
The other two enforce distribution similarity between the two modalities.
A cross-modal embedding similarity loss $\mathcal{L}_{\text{E}}$
enforces both text $z^T$ and motion $z^M$ latent codes to be similar to each other (with a smooth L1 loss).
We set $\lambda_{\text{KL}}$ and $\lambda_{\text{E}}$ to $10^{-5}$ in our experiments
as in \cite{petrovich22temos}.

\noindent\textbf{Contrastive training}.
While TEMOS has a cross-modal embedding space,
its major drawback to be usable as an effective retrieval
model is that it is never trained with negatives, but only
positive motion-text pairs.
To overcome this limitation, we incorporate a contrastive
training with
the usage of negative samples to better structure the latent space. 
Given a batch of N (positive) pairs of latent codes $(z_1^T, z_1^M), \cdots, (z_N^T, z_N^M)$,
we define any pair $(z_i^T, z_j^M)$ with $i \neq j$ as negative. The similarity matrix
$S$ computes the pairwise cosine similarities for all pairs in the batch 
$S_{ij} = \text{cos}(z_i^T, z_j^M)$.
In contrast to Guo et al.~\cite{Guo_2022_CVPR},
who consider one random negative per batch (and use a margin loss),
we adopt the more recent formulation of InfoNCE~\cite{oord2018representation}, 
which was proven effective in other work \cite{clip-pmlr-v139-radford21a,Bain_2021_ICCV,li2022blip}.
This loss term can be defined as follows:
\begin{equation}
\small
\begin{split} 
    \mathcal{L}_{\text{NCE}} &= - \frac{1}{2N} \sum_{i} \left( \log \frac{\exp{S_{ii}/\tau}}{\sum_{j} \exp{S_{ij}/\tau}} + \log \frac{\exp{S_{ii}/\tau}}{\sum_{j} \exp{S_{ji}/\tau}} \right) ~,
\end{split}
\label{eq:infonce}
\end{equation}
where $\tau$ is the temperature hyperparameter.

\noindent\textbf{Training loss}. The total loss we use to train \methodshort{} 
is the weighted sum $ \mathcal{L}_{\text{TEMOS}} + \lambda_{\text{NCE}} \mathcal{L}_{\text{NCE}}$ 
where $\lambda_{\text{NCE}}$ controls the importance of the contrastive loss.

\subsection{Filtering negatives}
\label{subsec:filter}

As mentioned in Section~\ref{sec:intro},
the %
descriptions 
accompanying motion capture
collections can be repetitive or similar
across the training motions. We wish to prevent defining
negatives between text-motion pairs that
contain similar descriptions.

Consider for example the two text descriptions, 
``\textit{A human making a counterclockwise turn}'' and
``\textit{A person walks quarter a circle to the left}''.
In the KIT-ML benchmark~\cite{Plappert2016_KIT_ML}, these two descriptions
appear as two different annotations for the same motion. %
Due to the flexibility and  ambiguity of natural language,
different words may describe the same concepts
(e.g., `counterclockwise', `circle to the left'). %

During training, the random selection of batches can adversely 
affect the results because the model may have to push away 
two latent vectors that correspond to similar meanings. 
This can force the network to focus on unimportant details
(e.g., `someone' vs `human'),
ultimately resulting in decreased performance due to
unstable behavior and reduced robustness to text variations.

To alleviate this issue, we leverage an external
large language model to provide sentence
similarity scores. In particular, we
use MPNet~\cite{mpnet2020song}
to encode sentences and compute similarities between
two text descriptions.
We then determine whether to filter a pair
of text descriptions ($t_1$, $t_2$) 
if their similarity is higher than a certain threshold,
referring them as `wrong negatives'.
During training, we filter wrong negative pairs from the loss computation.
Note that we refrain from defining them as positives, as the language model
may also incorrectly say two descriptions are similar when they are not. %

\subsection{Implementation details}
\label{subsec:implem_details}

We use the AdamW optimizer~\cite{adamw2019} with a learning rate
of $10^{-4}$ and a batch size of 32. Since the batch size
can be an important hyperparameter for the InfoNCE loss,
due to determining the number of negatives,
we report experimental results with different values.
The latent dimensionality of the embeddings is $d=256$.
We set
the temperature $\tau$ to $0.1$, and the weight of the contrastive loss term
 $\lambda_{\text{NCE}}$ to $0.1$.
 The threshold to filter negatives is set to 0.8.
 We provide experimental analyses to measure the sensitivity to these added hyperparameters.

\section{Experiments}
\label{sec:experiments}

We start by describing the datasets and evaluation protocol
used in the experiments (Section~\ref{subsec:datasets}).
We then report the performance of our model
on our new retrieval benchmark along with comparison to
prior work (Section~\ref{subsec:sota}).
Next, we present our ablation study measuring
the effects of the additional contrastive loss,
the negative filtering, and the hyperparameters (Section~\ref{subsec:ablations}).
Finally, we provide qualitative results for
retrieval (Section~\ref{subsec:qualitative}), and
\if\withaction1{our use cases of zero-shot action recognition
and moment retrieval (Section~\ref{subsec:usecases}).
} \else{our use case of moment retrieval (Section~\ref{subsec:usecases}).
} \fi

\begin{table*}
    \centering
    \setlength{\tabcolsep}{4pt}
    \resizebox{0.99\linewidth}{!}{
    \begin{tabular}{ll|cccccc|cccccc}
        \toprule
         \multirow{2}{*}{\textbf{Protocol}} & \multirow{2}{*}{\textbf{Methods}} & \multicolumn{6}{c|}{Text-motion retrieval} & \multicolumn{6}{c}{Motion-text retrieval} \\
         & & \small{R@1 $\uparrow$} & \small{R@2 $\uparrow$} & \small{R@3 $\uparrow$} &  \small{R@5 $\uparrow$} & \small{R@10 $\uparrow$} & \small{MedR $\downarrow$} & \small{R@1 $\uparrow$} & \small{R@2 $\uparrow$} & \small{R@3 $\uparrow$} & \small{R@5 $\uparrow$} & \small{R@10 $\uparrow$} & \small{MedR $\downarrow$} \\
\midrule
(a) All & \textbf{\texttt{TEMOS}~\cite{petrovich22temos}} &  2.12 &  4.09 &  5.87 &  8.26 & 13.52 & 173.0 &  3.86 &  4.54 &  6.94 &  9.38 & 14.00 & 183.25 \\
             &        \textbf{Guo et al.~\cite{Guo_2022_CVPR}} &  1.80 &  3.42 &  4.79 &  7.12 & 12.47 & 81.00 &  2.92 &  3.74 &  6.00 &  8.36 & 12.95 & 81.50 \\
             &                         \textbf{\methodshort{}} &  \textbf{5.68} & \textbf{10.59} & \textbf{14.04} & \textbf{20.34} & \textbf{30.94} & \textbf{28.00} &  \textbf{9.95} & \textbf{12.44} & \textbf{17.95} & \textbf{23.56} & \textbf{32.69} & \textbf{28.50} \\
\midrule
(b) All with threshold & \textbf{\texttt{TEMOS}~\cite{petrovich22temos}} &  5.21 &  8.22 & 11.14 & 15.09 & 22.12 & 79.00 &  5.48 &  6.19 &  9.00 & 12.01 & 17.10 & 129.0 \\
                            &        \textbf{Guo et al.~\cite{Guo_2022_CVPR}} &  5.30 &  7.83 & 10.75 & 14.59 & 22.51 & 54.00 &  4.95 &  5.68 &  8.93 & 11.64 & 16.94 & 69.50 \\
                            &                         \textbf{\methodshort{}} & \textbf{11.60} & \textbf{15.39} & \textbf{20.50} & \textbf{27.72} & \textbf{38.52} & \textbf{19.00} & \textbf{13.20} & \textbf{15.73} & \textbf{22.03} & \textbf{27.65} & \textbf{37.63} & \textbf{21.50} \\
\midrule
(c) Dissimilar subset & \textbf{\texttt{TEMOS}~\cite{petrovich22temos}} & 33.00 & 42.00 & 49.00 & 57.00 & 66.00 &  4.00 & 35.00 & 44.00 & 50.00 & 56.00 & 70.00 &  3.50 \\
                      &        \textbf{Guo et al.~\cite{Guo_2022_CVPR}} & 34.00 & 48.00 & 57.00 & 72.00 & 84.00 &  3.00 & 34.00 & 47.00 & 59.00 & 72.00 & 83.00 &  3.00 \\
                      &                         \textbf{\methodshort{}} & \textbf{47.00} & \textbf{61.00} & \textbf{71.00} & \textbf{80.00} & \textbf{86.00} &  \textbf{2.00} & \textbf{48.00} & \textbf{63.00} & \textbf{69.00} & \textbf{80.00} & \textbf{84.00} &  \textbf{2.00} \\
\midrule
(d) Small batches \cite{Guo_2022_CVPR} & \textbf{\texttt{TEMOS}~\cite{petrovich22temos}} & 40.49 & 53.52 & 61.14 & 70.96 & 84.15 &  2.33 & 39.96 & 53.49 & 61.79 & 72.40 & 85.89 &  2.33 \\
                                       &        \textbf{Guo et al.~\cite{Guo_2022_CVPR}} & 52.48 & 71.05 & 80.65 & 89.66 & \textbf{96.58} &  1.39 & 52.00 & 71.21 & 81.11 & 89.87 & \textbf{96.78} &  1.38 \\
                                       &                         \textbf{\methodshort{}} & \textbf{67.16} & \textbf{81.32} & \textbf{86.81} & \textbf{91.43} & 95.36 &  \textbf{1.04} & \textbf{67.97} & \textbf{81.20} & \textbf{86.35} & \textbf{91.70} & 95.27 &  \textbf{1.03} \\
    \bottomrule        
    \end{tabular}
    }
    \vspace{0.05in}
    \caption{\textbf{Text-to-motion retrieval benchmark on HumanML3D:}
    We establish four evaluation protocols as described in
    Section~\ref{subsec:datasets}, with decreasing difficulty from (a) to
    (d). Our model \methodshort{} substantially outperforms the prior
    work of Guo et al.~\cite{Guo_2022_CVPR} and \texttt{TEMOS}~\cite{petrovich22temos}, on the challenging H3D dataset.
    }
    \label{tab:benchmark_h3d}
\end{table*}

\begin{table*}
    \centering
    \setlength{\tabcolsep}{4pt}
    \resizebox{0.99\linewidth}{!}{
    \begin{tabular}{ll|cccccc|cccccc}
        \toprule
         \multirow{2}{*}{\textbf{Protocol}} & \multirow{2}{*}{\textbf{Methods}} & \multicolumn{6}{c|}{Text-motion retrieval} & \multicolumn{6}{c}{Motion-text retrieval} \\
         & & \small{R@1 $\uparrow$} & \small{R@2 $\uparrow$} & \small{R@3 $\uparrow$} & \small{R@5 $\uparrow$} & \small{R@10 $\uparrow$} & \small{MedR $\downarrow$} & \small{R@1 $\uparrow$} & \small{R@2 $\uparrow$} & \small{R@3 $\uparrow$} & \small{R@5 $\uparrow$} & \small{R@10 $\uparrow$} & \small{MedR $\downarrow$} \\
\midrule
(a) All & \textbf{\texttt{TEMOS}~\cite{petrovich22temos}} &  7.11 & 13.25 & 17.59 & 24.10 & 35.66 & 24.00 & \textbf{11.69} & \textbf{15.30} & \textbf{20.12} & 26.63 & 36.39 & 26.50 \\
             &        \textbf{Guo et al.~\cite{Guo_2022_CVPR}} &  3.37 &  6.99 & 10.84 & 16.87 & 27.71 & 28.00 &  4.94 &  6.51 & 10.72 & 16.14 & 25.30 & 28.50 \\
             &                         \textbf{\methodshort{}} &  \textbf{7.23} & \textbf{13.98} & \textbf{20.36} & \textbf{28.31} & \textbf{40.12} & \textbf{17.00} & 11.20 & 13.86 & \textbf{20.12} & \textbf{28.07} & \textbf{38.55} & \textbf{18.00} \\
\midrule
(b) All with threshold & \textbf{\texttt{TEMOS}~\cite{petrovich22temos}} & 18.55 & 24.34 & 30.84 & 42.29 & 56.39 &  7.00 & 17.71 & 22.41 & 28.80 & 35.42 & 47.11 & 13.25 \\
                            &        \textbf{Guo et al.~\cite{Guo_2022_CVPR}} & 13.25 & 22.65 & 29.76 & 39.04 & 49.52 & 11.00 & 10.48 & 13.98 & 20.48 & 27.95 & 38.55 & 17.25 \\
                            &                         \textbf{\methodshort{}} & \textbf{24.58} & \textbf{30.24} & \textbf{41.93} & \textbf{50.48} & \textbf{60.36} &  \textbf{5.00} & \textbf{19.64} & \textbf{23.73} & \textbf{32.53} & \textbf{41.20} & \textbf{53.01} &  \textbf{9.50} \\
\midrule
(c) Dissimilar subset & \textbf{\texttt{TEMOS}~\cite{petrovich22temos}} & 24.00 & 40.00 & 46.00 & 54.00 & 70.00 &  5.00 & 33.00 & 39.00 & 45.00 & 49.00 & 64.00 &  6.50 \\
                      &        \textbf{Guo et al.~\cite{Guo_2022_CVPR}} & 16.00 & 29.00 & 36.00 & 48.00 & 66.00 &  6.00 & 24.00 & 29.00 & 36.00 & 46.00 & 66.00 &  7.00 \\
                      &                         \textbf{\methodshort{}} & \textbf{26.00} & \textbf{46.00} & \textbf{60.00} & \textbf{70.00} & \textbf{83.00} &  \textbf{3.00} & \textbf{34.00} & \textbf{45.00} & \textbf{60.00} & \textbf{69.00} & \textbf{82.00} &  \textbf{3.50} \\
\midrule
(d) Small batches \cite{Guo_2022_CVPR} & \textbf{\texttt{TEMOS}~\cite{petrovich22temos}} & 43.88 & 58.25 & 67.00 & 74.00 & 84.75 &  2.06 & 41.88 & 55.88 & 65.62 & 75.25 & 85.75 &  2.25 \\
                                       &        \textbf{Guo et al.~\cite{Guo_2022_CVPR}} & 42.25 & 62.62 & 75.12 & 87.50 & \textbf{96.12} &  1.88 & 39.75 & 62.75 & 73.62 & 86.88 & \textbf{95.88} &  1.95 \\
                                       &                         \textbf{\methodshort{}} & \textbf{49.25} & \textbf{69.75} & \textbf{78.25} & \textbf{87.88} & 95.00 &  \textbf{1.50} & \textbf{50.12} & \textbf{67.12} & \textbf{76.88} & \textbf{88.88} & 94.75 &  \textbf{1.53} \\
\bottomrule        
    \end{tabular}
    }
    \vspace{0.05in}
    \caption{\textbf{Text-to-motion retrieval benchmark on KIT-ML:}
    As in Table~\ref{tab:benchmark_h3d}, we report the four evaluation protocols,
    this time on the KIT dataset.
    Again, \methodshort{} significantly improves over 
    Guo et al.~\cite{Guo_2022_CVPR} and \texttt{TEMOS}~\cite{petrovich22temos} across all protocols and metrics.
    }
    \label{tab:benchmark_kitml}
\end{table*}

\subsection{Datasets and evaluation}
\label{subsec:datasets}

\paragraph{HumanML3D dataset (H3D)~\cite{Guo_2022_CVPR}}
provides natural language labels to
describe the motions in AMASS~\cite{AMASS:ICCV:2019} and HumanAct12~\cite{chuan2020action2motion} motion capture collections.
We follow the motion pre-processing procedure of \cite{Guo_2022_CVPR},
and apply the SMPL layer~\cite{smpl2015} to extract joint positions,
canonicalize the skeletons to share the same topology (i.e., same bone lengths),
then compute motion features (extracting local positions, velocities and foot contacts similar to Holden~et~al.~\cite{Holden2017PhasefunctionedNN}). 
The data is then augmented by mirroring left and right (both in motions
and their corresponding texts).
After this procedure, and following the official split, we obtain $23384$, $1460$, $4380$ motions for the training, validation, and test sets, respectively. %
On average, each motion is annotated 3.0 times with different text.
During training we randomly select one as the matching text,
for testing we use the first text.

\noindent\textbf{KIT Motion-Language dataset (KIT)~\cite{Plappert2016_KIT_ML}}
also come from motion capture data, with an emphasis on locomotion motions.
It originally consists of 
3911 motion sequences and 6278 text sentences. 
We pre-process the motions with the same procedure as in H3D.
The data is split into $4888$, $300$, $830$ motions for training, validation, and test sets, respectively. In this dataset, each motion is annotated 2.1 times on average.

\noindent\textbf{Evaluation protocol.}
We report standard retrieval performance measures:
recall at several ranks, R@1, R@2, etc.\ for both
text-to-motion and motion-to-text tasks 
\cam{(identical to the R Precision metrics adopted by Guo et al.~\cite{Guo_2022_CVPR})}. 
Recall at rank $k$ measures the percentage of times the correct
label is among the top $k$ results; therefore higher is better. 
We additionally report median rank (MedR), where lower is better. 
\cam{Note that the \textit{retrieval} performance is evaluated on a gallery of unseen real motions (i.e., test set). A study of the \textit{synthesis} performance of \methodshort{} can be found in \if\sepappendix1{Appendix~B.}
\else{Appendix~\ref{appendix:experiments}.}
\fi}

We define several evaluation protocols, mainly
changing the gallery set.
(a) \textbf{All} the test set is used as a first protocol,
without any modification. This set is partially problematic
because there are repetitive texts across motions, or just minor differences
(e.g., person vs human, walk vs walking).
(b) \textbf{All with threshold} means we search over all the test set, but, in this case, we 
accept a retrieved motion as correct if its text label
is similar to the query text above a threshold.
For example, retrieving the motion corresponding to \textit{``A human walks forward''} should be correct when the input query is \textit{``Someone is walking forward''}.
We set a high threshold of 0.95 (scaled between [0, 1]) %
to remove very similar texts without removing too many fine-grained details. 
\if\sepappendix1{Appendix~A}
\else{Appendix~\ref{appendix:stats}}
\fi
provides statistics on how often similar text descriptions appear in the datasets.
(c) \textbf{Dissimilar subset} refers to sampling 100 motion-text pairs whose
texts are maximally far from each other (using
an approximation of the quadratic knapsack problem \cite{aiderhal04011129}
).
This evaluation can be considered
as an easy, but clean, subset of the previous ones.
(d) \textbf{Small batches} is included to mimic the protocol
described by Guo~et~al.~\cite{Guo_2022_CVPR}, who randomly pick
batches of 32 motion-text pairs and report the average performance.
An ideal evaluation metric should not have randomness,
and a gallery size of 32 is relatively easy compared to the previous protocols.

\subsection{A new benchmark \& comparison to prior work}
\label{subsec:sota}
We present the performance
of our model on this new retrieval benchmark, on 
H3D (Table~\ref{tab:benchmark_h3d})
and
KIT (Table~\ref{tab:benchmark_kitml})
datasets, across all evaluation protocols.
We also compare against prior work
TEMOS~\cite{petrovich22temos} and
Guo~et~al.~\cite{Guo_2022_CVPR}.
For TEMOS, we retrain their model on both datasets
to have a comparable benchmark since the original model
differs in motion representation and lacks left/right data augmentation 
(and they only provide a KIT-pretrained model, not H3D).
For \cite{Guo_2022_CVPR},
we take their publicly available models trained on these two datasets.

TEMOS in particular is not designed to perform well
on retrieval, since its cross-modal embedding space
is only trained with positive pairs.
However, Guo~et~al.\ train contrastively
with negatives as well, using a margin loss \cite{hadsell2006}.
For all 4 evaluation sets with varying difficulties,
\methodshort{} outperforms the prior work, suggesting
our model better captures the finegrained nature of motion
descriptions.
The model of \cite{Guo_2022_CVPR} is adopted as part of motion synthesis evaluation in several works. \methodshort{} may therefore
provide a better alternative.
Our significant improvements
over the state of the art
can be credited to (i) jointly training for
synthesis and retrieval, (ii) adopting the more
recent contrastive objective, InfoNCE~\cite{oord2018representation},
while (iii) carefully eliminating
wrong negatives.
In the following, we ablate these
components in controlled experiments.

\subsection{Ablation study}
\label{subsec:ablations}

The rest of the quantitative evaluation
uses the `(b) All with threshold'
evaluation protocol, on the KIT dataset.

\begin{table}
    \centering
    \setlength{\tabcolsep}{4pt}
    \resizebox{0.99\linewidth}{!}{
    \begin{tabular}{c|cc|cccc|cccc}
        \toprule
            Motion & \multirow{2}{*}{\textbf{InfoNCE}} & \multirow{2}{*}{\textbf{Margin}} & \multicolumn{4}{c|}{Text-motion retrieval} & \multicolumn{4}{c}{Motion-text retrieval} \\
          \textbf{Recons.} & & & \small{R@1 $\uparrow$} & \small{R@2 $\uparrow$} & \small{R@3 $\uparrow$}  & \small{MedR $\downarrow$} & \small{R@1 $\uparrow$} & \small{R@2 $\uparrow$} & \small{R@3 $\uparrow$} & \small{MedR $\downarrow$} \\
\midrule
\xmark & \xmark & \cmark & 15.06 & 22.17 & 25.78 & 12.00 &  8.19 & 11.57 & 16.39 & 19.50 \\
\xmark & \cmark & \xmark & 19.76 & 25.30 & 36.87 &  6.00 & 17.47 & 19.76 & 30.60 &  \textbf{9.50} \\
\midrule
\cmark & \xmark & \xmark & 18.55 & 24.34 & 30.84 &  7.00 & 17.71 & 22.41 & 28.80 & 13.25 \\
\midrule
\cmark & \xmark & \cmark & 19.88 & 24.46 & 34.46 &  7.00 & 14.70 & 19.76 & 28.19 & 12.50 \\
\cmark & \cmark & \xmark & \textbf{24.58} & \textbf{30.24} & \textbf{41.93} &  \textbf{5.00} & \textbf{19.64} & \textbf{23.73} & \textbf{32.53} &  \textbf{9.50} \\
\bottomrule        
    \end{tabular}   
    }
    \vspace{0.05in}
    \caption{\textbf{Losses:}
    We experiment with various loss definitions (i) with/without the motion
    reconstruction, and (ii) the choice of the contrastive loss between InfoNCE and margin-based. We see that InfoNCE~\cite{oord2018representation}
    is a better
    alternative to the contrastive loss with Euclidean margin~\cite{hadsell2006} (employed
    by Guo~et~al.~\cite{Guo_2022_CVPR}).
    The reconstruction loss through the motion decoder branch further boosts
    the results.
    }
    \vspace{-0.3cm}
    \label{tab:loss_ablation_kitml}
\end{table}

\noindent\textbf{Which losses matter?}
Table~\ref{tab:loss_ablation_kitml} compares several variants of~\methodshort{} where we check (a) whether the jointly trained motion
synthesis branch helps retrieval, and (b) how important the form of
the contrastive loss is.
When removing the synthesis branch and only using the %
the contrastive loss,
we perform a deterministic encoding
(i.e., with a single token instead of two tokens $\mu$, $\sigma$).
First, we see that the motion synthesis branch certainly improves results compared with only training using a contrastive loss
(e.g., 41.93 vs 36.87 R@3).
This possibly forces the latent vector to capture the full
content of the input text (i.e., instead of picking up on a subset of words,
or bag-of-words \cite{doven2023,yuksekgonul2022}
upon finding a shortcut that satisfies the contrastive loss).
Second, in the presence of a contrastive loss,
the InfoNCE formulation is significantly better than the margin loss employed
by previous work \cite{Guo_2022_CVPR} (41.93 vs 34.46 R@1).
Note that for this experiment, we keep the same
negative filtering for both the margin loss and InfoNCE \cam{(we provide additional experiments without the negative filtering in \if\sepappendix1{Appendix~B).}
\else{Appendix~\ref{appendix:experiments}).}
\fi}

\begin{table}
    \centering
    \setlength{\tabcolsep}{4pt}
    \resizebox{0.99\linewidth}{!}{
    \begin{tabular}{c|cccc|cccc}
        \toprule
            \multirow{2}{*}{\textbf{Threshold}} & \multicolumn{4}{c|}{Text-motion retrieval} & \multicolumn{4}{c}{Motion-text retrieval} \\
          & \small{R@1 $\uparrow$} & \small{R@2 $\uparrow$} & \small{R@3 $\uparrow$}  & \small{MedR $\downarrow$} & \small{R@1 $\uparrow$} & \small{R@2 $\uparrow$} & \small{R@3 $\uparrow$} & \small{MedR $\downarrow$} \\
\midrule
  0.55 & 19.40 & 23.25 & 30.48 &  9.00 & 17.83 & 21.69 & 29.52 & 14.00 \\
   0.60 & 17.95 & 26.87 & 36.87 &  6.00 & 20.60 & 24.70 & 31.81 & 11.25 \\
  0.65 & 23.01 & 28.67 & 36.39 &  7.00 & 19.04 & 21.69 & 29.76 & 11.50 \\
   0.70 & 22.29 & 29.64 & 38.80 &  6.00 & 18.19 & 22.77 & 32.05 &  \textbf{9.00} \\
  0.75 & 20.00 & 27.11 & 37.83 &  6.00 & 20.24 & 24.46 & \textbf{34.22} &  9.50 \\
   0.80 & \textbf{24.58} & \textbf{30.24} & \textbf{41.93} &  \textbf{5.00} & 19.64 & 23.73 & 32.53 &  9.50 \\
  0.85 & 21.45 & 25.78 & 38.43 &  6.50 & \textbf{20.84} & 24.10 & 33.37 &  9.50 \\
   0.90 & 23.25 & 30.12 & 40.48 &  6.00 & 20.00 & \textbf{25.18} & 33.13 &  9.50 \\
  0.95 & 20.48 & 26.99 & 38.43 &  6.00 & 19.28 & 23.37 & 31.93 & 10.25 \\
\xmark & 22.17 & 27.83 & 36.02 &  7.00 & 16.75 & 21.33 & 32.17 & 11.50 \\
\bottomrule        
    \end{tabular}
    }
    \vspace{0.05in}
    \caption{\textbf{Filtering negatives:}
    We compare several threshold values for filtering negatives
    from the loss comparison due to having similar texts.
    We observe that removing negatives 
    based on text similarity above 0.8 (from a scale between [0,1]) %
    performs well overall.
    }
    \vspace{-0.4cm}
    \label{tab:filtering_ablation_kitml}
\end{table}

\noindent\textbf{The effect of filtering negatives.}
As explained in Section~\ref{subsec:filter},
during training we filter out pairs whose
texts are closer than a threshold in an embedding space,
and do not count them in the contrastive loss computation.
Note that we still keep each item in the batch for
the motion synthesis objective.
In Table~\ref{tab:filtering_ablation_kitml},
we perform experiments with a range of different values
for this threshold selection.
On a scale between [0, 1], a threshold of 0.8
has the best results, balancing
keeping a sufficient number of negatives and removing
the wrong ones. Without filtering at all, the performance
remains at 36.02 R@3 (compared to 41.93).
We provide statistics on the percentage of filtered pairs in
\if\sepappendix1{Appendix~A.}
\else{Appendix~\ref{appendix:stats}.}
\fi

\noindent\textbf{Hyperparameters of the contrastive training.}
We show the sensitivity of our model to several hyperparameters
added when extending TEMOS:
(i) temperature $\tau$ of the cross entropy of InfoNCE~\cite{oord2018representation} in Eq.\ref{eq:infonce},
(ii) the $\lambda_{\text{NCE}}$ weighting parameter, and
(iii) the batch size, which determines the number of negatives.
We see in Table~\ref{tab:cparam_ablation_kitml} that
the model is indeed sensitive to the temperature,
which is a common observation in other settings.
The weight parameter and the batch size are relatively
less important while also influencing the results
to a certain extent.
An experiment with the latent dimensionality 
hyperparameter
can be found in
\if\sepappendix1{Appendix~B.}
\else{Appendix~\ref{appendix:experiments}.}
\fi

\begin{table}
    \centering
    \setlength{\tabcolsep}{4pt}
    \resizebox{0.99\linewidth}{!}{
    \begin{tabular}{c|cccc|cccc}
        \toprule
            \textbf{Temp.} &  \multicolumn{4}{c|}{Text-motion retrieval} & \multicolumn{4}{c}{Motion-text retrieval} \\
          \textbf{$\tau$}& \small{R@1 $\uparrow$} & \small{R@2 $\uparrow$} & \small{R@3 $\uparrow$}  & \small{MedR $\downarrow$} & \small{R@1 $\uparrow$} & \small{R@2 $\uparrow$} & \small{R@3 $\uparrow$} & \small{MedR $\downarrow$} \\
\midrule
0.001 &  9.52 & 21.81 & 27.23 & 12.00 &  7.47 &  9.76 & 16.51 & 15.50 \\
 0.01 & 21.45 & 29.04 & 38.80 &  6.00 & \textbf{21.08} & \textbf{27.11} & \textbf{33.61} &  \textbf{9.50} \\
  0.1 & \textbf{24.58} & \textbf{30.24} & \textbf{41.93} &  \textbf{5.00} & 19.64 & 23.73 & 32.53 &  \textbf{9.50} \\
  1.0 &  1.08 &  1.93 &  3.61 & 306.5 &  1.81 &  1.93 &  2.41 & 372.0 \\
\bottomrule
\multicolumn{9}{c}{(a)}\\
\multicolumn{9}{c}{}\\
\toprule
            \textbf{Weight} & \multicolumn{4}{c|}{Text-motion retrieval} & \multicolumn{4}{c}{Motion-text retrieval} \\
          \textbf{$\lambda_{\text{NCE}}$} & \small{R@1 $\uparrow$} & \small{R@2 $\uparrow$} & \small{R@3 $\uparrow$}  & \small{MedR $\downarrow$} & \small{R@1 $\uparrow$} & \small{R@2 $\uparrow$} & \small{R@3 $\uparrow$} & \small{MedR $\downarrow$} \\
\midrule
0.001 & 18.55 & 23.25 & 36.75 &  7.00 & 18.19 & \textbf{24.34} & 31.45 & 11.50 \\
 0.01 & 20.84 & 26.99 & 37.23 &  7.00 & 18.92 & 23.13 & 32.17 & 10.25 \\
  0.1 & \textbf{24.58} & \textbf{30.24} & \textbf{41.93} &  \textbf{5.00} & \textbf{19.64} & 23.73 & 32.53 &  \textbf{9.50} \\
  1.0 & 19.52 & 24.46 & 34.46 &  7.00 & 19.04 & \textbf{24.34} & \textbf{35.06} &  \textbf{9.50} \\
\bottomrule        
\multicolumn{9}{c}{(b)}\\
\multicolumn{9}{c}{}\\
\toprule
            \textbf{Batch} & \multicolumn{4}{c|}{Text-motion retrieval} & \multicolumn{4}{c}{Motion-text retrieval} \\
          \textbf{size} & \small{R@1 $\uparrow$} & \small{R@2 $\uparrow$} & \small{R@3 $\uparrow$}  & \small{MedR $\downarrow$} & \small{R@1 $\uparrow$} & \small{R@2 $\uparrow$} & \small{R@3 $\uparrow$} & \small{MedR $\downarrow$} \\
\midrule
 16 & \textbf{25.42} & \textbf{31.57} & 40.12 &  6.00 & \textbf{20.36} & 24.10 & \textbf{33.73} &  \textbf{8.00} \\
 32 & 24.58 & 30.24 & \textbf{41.93} &  \textbf{5.00} & 19.64 & 23.73 & 32.53 &  9.50 \\
 64 & 20.24 & 26.51 & 38.19 &  6.00 & 19.52 & \textbf{24.22} & 32.05 &  9.50 \\
128 & 18.55 & 28.80 & 36.75 &  7.00 & 14.94 & 18.43 & 26.14 & 11.50 \\
\bottomrule
\multicolumn{9}{c}{(c)}\\
    \end{tabular}
    }
    \caption{\textbf{Hyperparameters of the contrastive training:}
    We measure the sensitivity to the parameters $\tau$ (temperature), $\lambda_c$ the weight of the contrastive loss, and the batch size.
    Note that the learning rate is proportionally altered when changing the batch size. \cam{We display a wide range of values to show the full trends.}
    }
    \label{tab:cparam_ablation_kitml}
\end{table}

\begin{figure*}
    \includegraphics[width=0.96\linewidth]{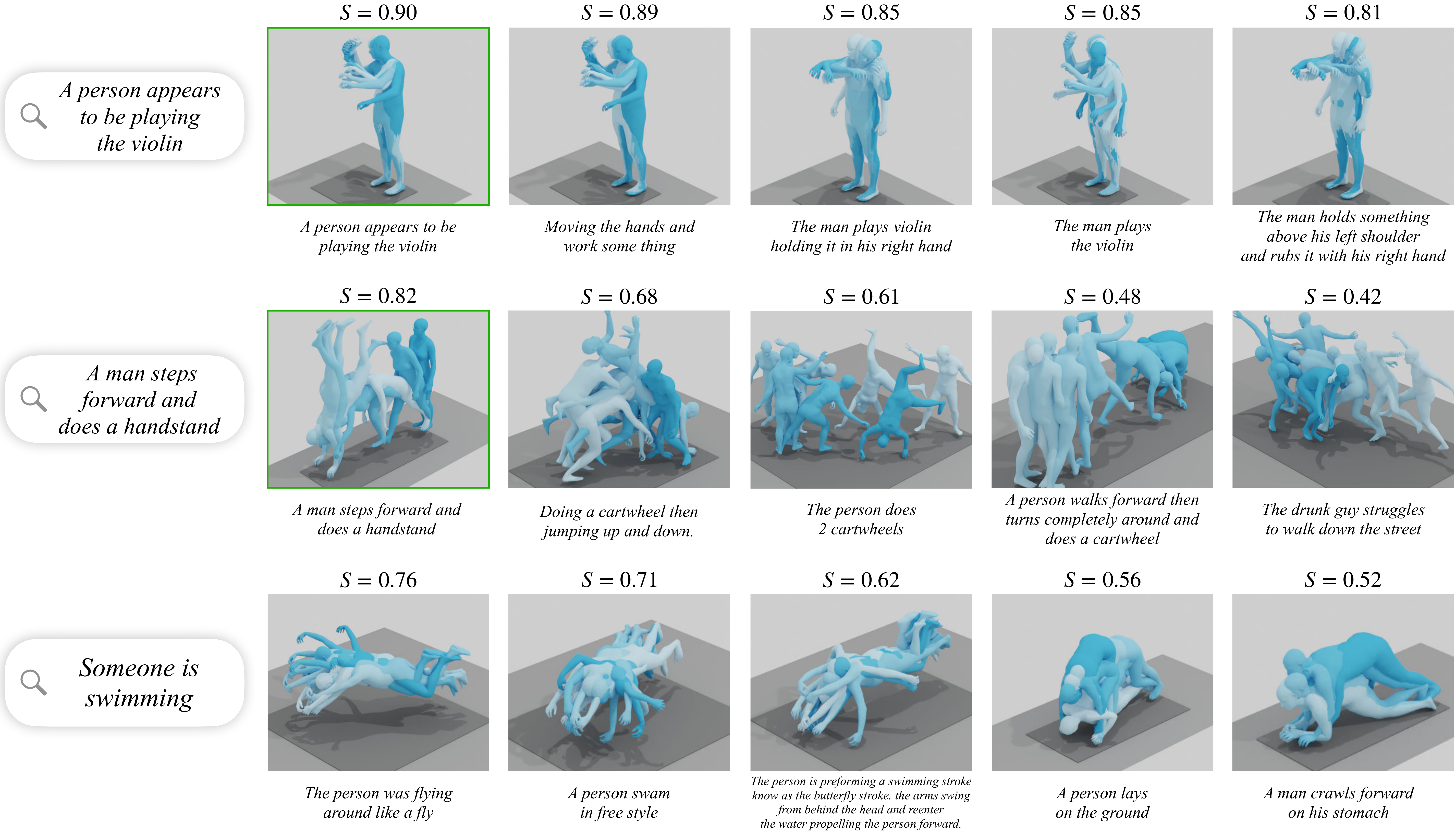}
    \caption{\textbf{Qualitative retrieval results:}
        We demonstrate example queries on the left, and
        corresponding retrieved motions on the right, ranked
        by text-motion similarity. The similarity values are
        displayed on the top. For each retrieved motion, we also
        show their accompanying ground-truth text label; note that
        we do not use these descriptions, but only provide them
        for analysis purposes. 
        The motions from the gallery are all from the test set (unseen during training).
        In the first row, all top-5 retrieved motions correspond visually to `playing violin' and the similarity scores are high $>0.80$. 
        In the second row, we correctly retrieve the `handstand' motion at top-1, but the other motions mainly perform `cartwheel' (which involves shortly standing on hands), but with a lower similarity score $<0.70$.        
    For the last example, we query a free-form text `Someone is swimming', which does not exist in the gallery (but the word `swim' does). The model successfully finds swimming motions among the top-3, and the other two motions involve the body parallel to the ground.
    }
    \vspace{-0.4cm}
    \label{fig:qualitative}
\end{figure*}

\subsection{Qualitative results}
\label{subsec:qualitative}
In Figure~\ref{fig:qualitative}, we provide
sample qualitative results for text-to-motion retrieval on the full test set of H3D.
For each query text displayed on the left, top-5
retrieved motions are shown on the right along with
their similarity scores.
Note that the ground-truth text labels (at the bottom of each motion) for the retrieved
motions are not used, and the gallery motions are unseen at training.
For the first two examples with `playing violin' and `handstand',
we retrieve the ground-truth motion at rank 1. We observe that
the next ranked motions depict visually similar motions as well (e.g., `cartwheel' involves standing on the hands).
For the free-from prompt example `Someone is swimming' (i.e., the exact text does not appear in the gallery), the three first motions resemble or involve the swimming action,
whereas motions at ranks 4 and 5 are incorrect. We notice that the incorrect motions have a low similarity ($<0.6$), and the human bodies are rotated similarly as in swimming.
More qualitative results can be seen in
\if\sepappendix1{Appendix~C,}
\else{Appendix~\ref{appendix:qualitative},}
\fi
as well as our supplementary video on the project page.

\begin{figure}
 \includegraphics[width=0.9\linewidth]{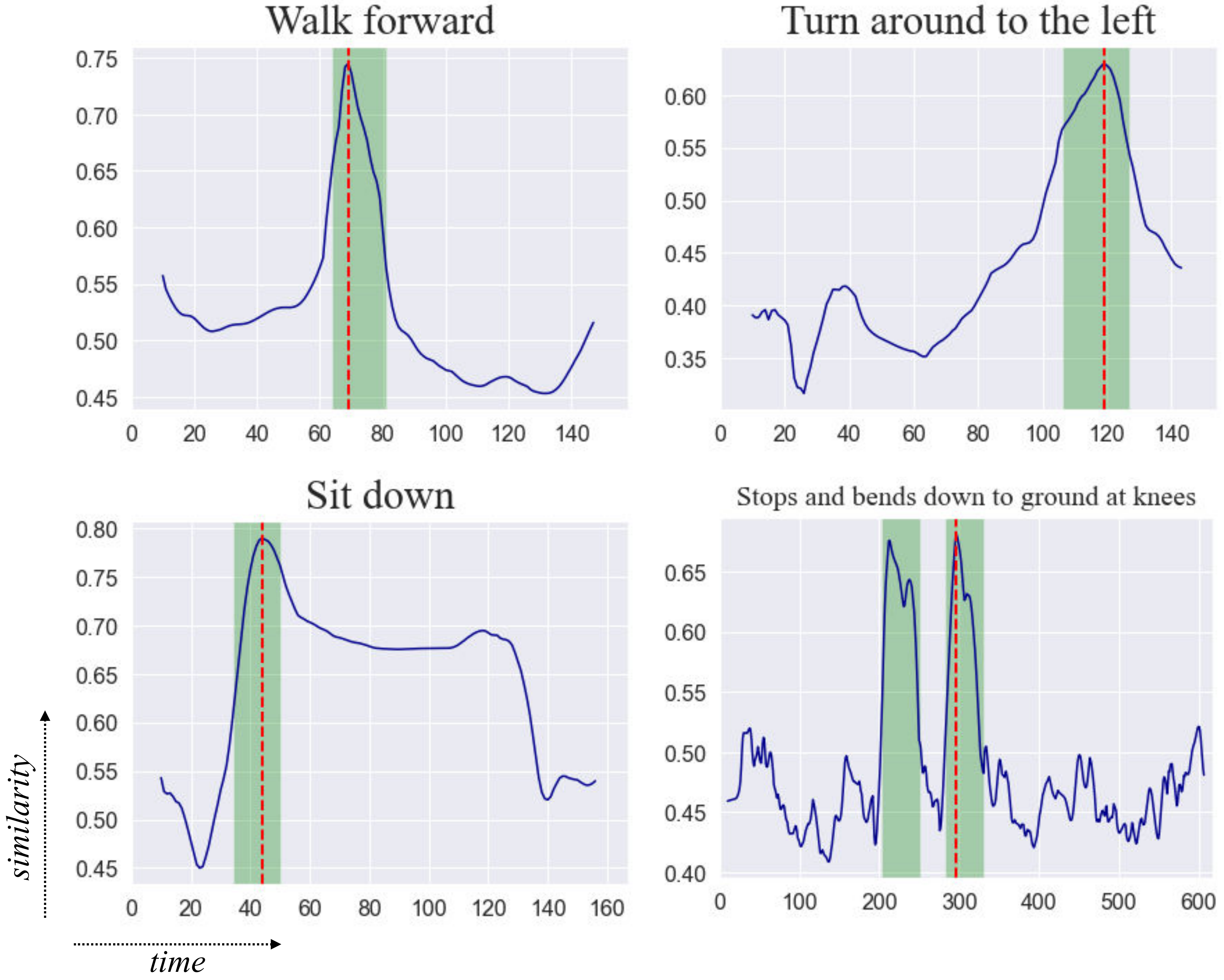}
	\caption{\textbf{Moment retrieval:}
            We plot the similarity between the temporally
            annotated BABEL text labels and the motions
            in a sliding window manner, and obtain a 1D
            signal over time (blue). We observe that a localization
            ability emerges from our model, even though it was not
            trained  for temporal localization,
            and was not with the domain of BABEL labels.
            The ground-truth temporal span is denoted in green and
            the maximum similarity
            is marked with a dashed red line.
            More examples are provided in
            \if\sepappendix1{Appendix Figure~A.2.}
            \else{Appendix Figure~\ref{fig:localization}.}
            \fi
	}
	\label{fig:usecases}
        \vspace{-0.4cm}
\end{figure}

\subsection{Use case: Moment retrieval}
\label{subsec:usecases}
While our focus is retrieval, once our model is trained,
it can be used for a different use case. Here, we
test the limits of our approach, by  qualitatively evaluating
the capability of \methodshort{} on the task of
temporally localizing a natural language query
in a long 3D motion sequence.
This is similar in spirit to moment retrieval
in videos~\cite{escorcia2019, tvr2020, hendricks2017, gao2017, regneri2013}.
It is also related to categorical action localization
in 3D motions~\cite{sun2022}; however, our input is
free-form text instead of symbolic action classes.

\if\withaction1{
\paragraph{Zero-shot 3D action recognition.}
\paragraph{Moment retrieval in 3D motions using natural language queries.}
}\fi

In Figure~\ref{fig:usecases}, we show four examples,
where we apply a model pre-trained on H3D on BABEL
sequences.
In each example, the queried text  is displayed
on the top. The x-axis denotes the frame number,
the green rectangle represents the ground-truth location
for the given action, and the dashed red line marks the localization
with the maximum similarity. We simply compute the motion features
in a sliding window manner. The similarity between the
text label and a 20-frame window centered at each frame is shown
in the y-axis as a 1D plot over time. Despite our model
not being trained for temporal localization,
we observe its grounding potential.
Moreover, there exists a domain gap between BABEL label used at test time and H3D used for training.
Quantitative evaluation and more qualitative
examples are included in
\if\sepappendix1{Appendix~B.}
\else{Appendix~\ref{appendix:experiments}.}
\fi

\subsection{Limitations}
\label{subsec:limitations}
Our model comes with some limitations.
Compared to the vast amount of data
(e.g., 400M images \cite{schuhmann2021laion})
in image-text collections used to train competitive foundation models, our
motion-text training data can be considered extremely small
(e.g., 23K motions in H3D).
The generalization performance of motion retrieval
models to in-the-wild motions is therefore limited.
Data augmentations such as altering text
can potentially help to a certain extent (cf.~\cite{SINC:2023});
however, more motion capture is still needed. 
Another limitation concerns the case where one wishes
to replace motion synthesis by retrieving a training motion.
In this use case, the model requires the full encoded database
(i.e., encoded training set)
to be stored in memory, which can be memory-inefficient.
Note that the encoded representation is considerably smaller than the original motion data.
This could be dealt with using a hierarchical structuring of the motions and language descriptions (e.g.~into upper-body motions, lower-body motions, sitting motions, standing motions, etc.).

\vspace{-0.5em}
\section{Conclusion}
\label{sec:conclusion}

In this paper, we focus on the relatively
little-studied problem of motion retrieval with
natural language queries.
We introduce \methodshort{}, a framework
to jointly train text-to-motion retrieval
and text-to-motion synthesis, with a special
attention to the definition of negatives,
taking into account the fine-grained nature of
motion-language databases.
We significantly improve over prior work,
and provide a series of experiments
highlighting the importance of each component.

Future work may consider incorporating a language synthesis
branch, along with the motion synthesis branch,
to build a symmetrical framework, which could
bring further benefits.
\medskip

{ \small %
\noindent\textbf{Acknowledgements.}
This work was granted access to
the HPC resources of IDRIS under the allocation 2022-AD011012129R2 made by GENCI.
GV
acknowledges the ANR project CorVis ANR-21-CE23-0003-01.
The authors would like to thank Lucas Ventura and Charles Raude.

\noindent\textbf{MJB Disclosure}: \url{https://files.is.tue.mpg.de/black/CoI_ICCV_2023.txt}
\par
}

{\small
\bibliographystyle{ieee_fullname}
\bibliography{7_references}
}
\bigskip
{\noindent \large \bf {APPENDIX}}\\
\renewcommand{\thefigure}{A.\arabic{figure}} %
\setcounter{figure}{0} 
\renewcommand{\thetable}{A.\arabic{table}}
\setcounter{table}{0} 

\appendix

As mentioned in the main text, this appendix includes 
statistical analysis (Section~\ref{appendix:stats}), 
additional experimental results (Section~\ref{appendix:experiments}), 
and further qualitative results (Section~\ref{appendix:qualitative}).

\noindent\textbf{Supplementary video.}
In addition to this appendix, we provide a video on our project page to allow viewing motions dynamically. 
In the video, we demonstrate qualitative results for text-to-motion retrieval 
on the two datasets 
KIT~\cite{Plappert2016_KIT_ML} and H3D~\cite{Guo_2022_CVPR}. Moreover, we illustrate
the use case 
of moment retrieval on BABEL~\cite{babel2021}.

\noindent\textbf{Code \& Demo.}
We further provide the source code for
training and evaluation, along with an interactive demo, which we make
publicly available.

\section{Statistics}
\label{appendix:stats}

\noindent\textbf{Number of similar text descriptions in the test set.} 
As mentioned in 
\if\sepappendix1{Section 4.1}
\else{Section~\ref{subsec:datasets}}
\fi
of the main paper,
the evaluation protocol (b)
marks retrieved motions as correct if their corresponding
text is similar to the queried text above a threshold of 0.95
(note that this threshold is different from the one used in training).
Here, we report the total number of pairs
that are above this threshold for each dataset.
For KIT, on the $830$ sequences of the test set, 
there are $344,035$ unique pairs of texts ($830 * 829 / 2$) 
from which $2,467$ of them are similar (about $0.7$\% of the data). 
For H3D, on the $4,380$ sequences of the test set, 
there are  $9,590,010$ unique pairs of texts ($4380 * 4380 / 2$) 
from which $6,017$ of them are similar (about $0.06$\% of the data).

\noindent\textbf{Percentage of filtered negatives per batch during training.} 
To complement 
\if\sepappendix1{Tables 4 and 5}
\else{Tables~\ref{tab:filtering_ablation_kitml} and \ref{tab:cparam_ablation_kitml}}
\fi
of the main paper,
in Table~\ref{tab:avg_negatives}, we compute the amount
of negatives that are filtered on average per batch, depending 
on the threshold and the batch size.
In our current setting, 17.29\% of 
the negatives are discarded. We see that this 
rate remains similar across batch sizes.

\section{Additional experimental results}
\label{appendix:experiments}

\noindent\textbf{Motion synthesis results.}
\cam{As mentioned in 
\if\sepappendix1{Section 4.1}
\else{Section~\ref{subsec:datasets}}
\fi
of the main paper, we evaluate the synthesis performance of \methodshort{}. 
In Table~\ref{tab:synthesis}, we compare the performance of \methodshort{}, TEMOS and 
Guo et al.~\cite{Guo_2022_CVPR} under various settings. %
See the caption for explanations and comments.
}

\begin{table}
    \centering
    \setlength{\tabcolsep}{4pt}
    \resizebox{0.99\linewidth}{!}{
    \begin{tabular}{l|ccccccccc}
        \toprule
        Threshold & 0.55 & 0.6 & 0.65 & 0.7 & 0.75 & \textbf{0.8} & 0.85 & 0.9 & 0.95 \\
        \midrule
        \% filtered negatives & 98.04 & 88.04 & 68.56 & 48.27 & 31.54 & \textbf{17.29} & 7.41 & 2.78 & 0.71 \\
        \bottomrule
    \end{tabular}
    }
    \\ \vspace{0.3cm}
    \resizebox{0.59\linewidth}{!}{
    \begin{tabular}{l|ccccc}
        \toprule
        Batch size                & 16 & \textbf{32} & 64 & 128 \\
        \midrule
        \% filtered negatives& 17.02 & \textbf{17.29} & 16.96 & 17.28\\
        \bottomrule        
    \end{tabular}
    }
    \vspace{0.3cm}
    \caption{\textbf{Percentage of filtered negatives per batch in KIT:}
    We compute the average percentage of negative pairs per batch that are discarded from the loss computation due to text similarity. The percentage decreases with higher thresholds as expected (top), but the batch size does not have a significant impact (bottom).
    }
    \label{tab:avg_negatives}
\end{table}

\begin{table}
    \centering
    \setlength{\tabcolsep}{4pt}
   \resizebox{\linewidth}{!}{
        \begin{tabular}{l|ccc|ccc}
            \toprule
            & \multicolumn{3}{c|}{KIT-ML} & \multicolumn{3}{c}{H3D}\\
            \diagbox{\textbf{Motions}}{\textbf{Eval}} & Guo Ret. & TEMOS & TMR & Guo Ret. & TEMOS & TMR \\
            \midrule
            Real motions & 42.25 & 44.88 & 49.25  & 52.41 & 42.33 & 67.16 \\
            \midrule
            Guo Syn.~\cite{Guo_2022_CVPR} & 36.88 & 47.00 & 48.38 & 45.80 & 37.73 & 55.38 \\
            TEMOS~\cite{petrovich22temos} & 43.88 & 90.50 & 76.88 & 40.76 & 79.71 & 72.38 \\
            \methodshort & 43.50 & 71.88 & 89.25  & 44.67 & 57.35 & 92.44 \\
            \bottomrule        
        \end{tabular}   
   }
   \vspace{0.05in}
   \caption{
        \small
        \textbf{Motion synthesis results:}
        We report R@1 text-to-motion retrieval performance of \emph{generated}
        motions by the synthesis method of Guo et al.~\cite{Guo_2022_CVPR} (Guo Syn.),
        TEMOS~\cite{petrovich22temos}, and our TMR synthesis branch, as well as the `Real motions',
        on both KIT-ML (left) and H3D (right) benchmarks. Rows are different
        motion generation methods, columns are different retrieval evaluation models:
        retrieval method of Guo et al.~\cite{Guo_2022_CVPR} (Guo Ret.), TEMOS, and our TMR retrieval branch.
        We use the protocol (d), i.e., 32 gallery size protocol from~\cite{Guo_2022_CVPR} .
        We make several observations:
        (i) TMR, when used for motion synthesis, performs better than or similar to Guo Syn.~\cite{Guo_2022_CVPR} 
        across all 3 retrieval evaluation models, showing we do not sacrifice synthesis performance.
        (ii) Evaluation with retrieval models that can also perform synthesis (TEMOS and TMR) favors motions generated by their own model.
        (iii) Certain numbers are better than Real motions, potentially
        because generations are sometimes more faithful to the input text, which
        may incompletely describe the real motion, or due to the bias mentioned in (ii).
    }
    \label{tab:synthesis}
\end{table}

\begin{table}
    \centering
    \setlength{\tabcolsep}{4pt}
    \resizebox{0.99\linewidth}{!}{
    \begin{tabular}{c|cccc|cccc}
        \toprule
            \textbf{Latent dim.} & \multicolumn{4}{c|}{Text-motion retrieval} & \multicolumn{4}{c}{Motion-text retrieval} \\
          $d$ & \small{R@1 $\uparrow$} & \small{R@2 $\uparrow$} & \small{R@3 $\uparrow$}  & \small{MedR $\downarrow$} & \small{R@1 $\uparrow$} & \small{R@2 $\uparrow$} & \small{R@3 $\uparrow$} & \small{MedR $\downarrow$} \\
\midrule
 64 & 18.80 & 28.67 & 38.43 &  6.00 & 18.07 & 21.81 & 31.45 &  9.50 \\
128 & \textbf{25.90} & \textbf{31.20} & 40.72 &  6.00 & \textbf{23.73} & \textbf{27.35} & \textbf{36.39} &  \textbf{9.25} \\
256 & 24.58 & 30.24 & \textbf{41.93} &  \textbf{5.00} & 19.64 & 23.73 & 32.53 &  9.50 \\
512 & 23.13 & 28.43 & 35.42 &  7.00 & 20.36 & 26.39 & 33.61 & 10.50 \\
\bottomrule        
    \end{tabular}
    }
    \vspace{0.05in}
    \caption{\textbf{Latent dimensionality:}
    We experiment with the embedding space dimensionality,
    and observe that $d=128$ performs overall best.
    However, in all other experiments, we use
    $d=256$ as in TEMOS.
    }
    \label{tab:latent_dim_kitml}
\end{table}

\noindent\textbf{Latent dimensionality.}
As stated in
\if\sepappendix1{Section 3.4}
\else{Section~\ref{subsec:implem_details}}
\fi
of the main paper,
the dimensionality of the latent space is set to
$d=256$ as in TEMOS~\cite{petrovich22temos}.
In Table~\ref{tab:latent_dim_kitml},
we experiment with this architectural design choice,
and observe that $d=128$ brings overall better performance.

\noindent\textbf{Contrastive-only baseline.} 
\cam{
As outlined in 
\if\sepappendix1{Section 4.3}
\else{Section~\ref{subsec:ablations}}
\fi
of the main paper,
we also experiment with the contrastive model without negative filtering, and present the results in Table~\ref{tab:loss_ablation_kitml_no_filtering}.
The negative filtering overall improves the results both with the contrastive-only model and with the added synthesis branch (\methodshort{}). 
We note that the added synthesis branch empirically improves the results consistently. Similar conclusions were already made by the text-to-image multi-modal models such as BLIP~\cite{li2022blip} and CoCa~\cite{yu2022coca} which improve performance over contrastive-only CLIP~\cite{clip-pmlr-v139-radford21a} by adding a text synthesis loss.
}

\begin{table}
    \centering
    \setlength{\tabcolsep}{4pt}
    \resizebox{0.8\linewidth}{!}{
        \begin{tabular}{lc|cccc}
            \toprule
            & NF & \small{R@1 $\uparrow$} & \small{R@2 $\uparrow$} & \small{R@3 $\uparrow$}  & \small{MedR $\downarrow$} \\
            \midrule 
            Contrastive-only & \xmark & 19.16 & \textbf{25.54} & 33.13 &  8.00 \\
            Contrastive-only & \cmark & \textbf{19.76} & 25.30 & \textbf{36.87} & \textbf{6.00 } \\
            \midrule
            \methodshort & \xmark & 22.17 & 27.83 & 36.02 &  7.00 \\
            \methodshort & \cmark & \textbf{24.58} & \textbf{30.24} & \textbf{41.93} & \textbf{5.00} \\
            \bottomrule        
        \end{tabular}   
    }
    \vspace{0.05in}
    \caption{
    \textbf{Contrastive-only without negative filtering:} We report text-to-motion retrieval results on KIT-ML to analyze the impact of negative filtering (NF) on the contrastive-only baseline. First row is the \cam{supplemental} result, the rest are from the main paper.
    }
    \label{tab:loss_ablation_kitml_no_filtering}
\end{table}

\noindent\textbf{Moment retrieval.} 
As presented in
\if\sepappendix1{Section 4.5}
\else{Section~\ref{subsec:usecases}}
\fi
of the main paper,
we localize a textual query within a motion, 
by computing the similarity between the text and several temporal crops of the motion in a zero-shot manner
(i.e., the model was not trained for this task,
nor has it seen BABEL texts).
Here, we provide additional qualitative results,
and also report quantitative metrics.

In Figure~\ref{fig:localization},
we provide complementary qualitative results to
\if\sepappendix1{Figure 4}
\else{Figure~\ref{fig:usecases}}
\fi
of the main paper.
At the right of Figure~\ref{fig:localization} (b), we
also show the localization potential on four very long sequences. As the search space gets larger,
the similarity plot gets noisier; however,
the maximum similarity still occurs at the
ground-truth location (marked in green).

\begin{figure}
	\centering
	\includegraphics[width=0.99\linewidth]{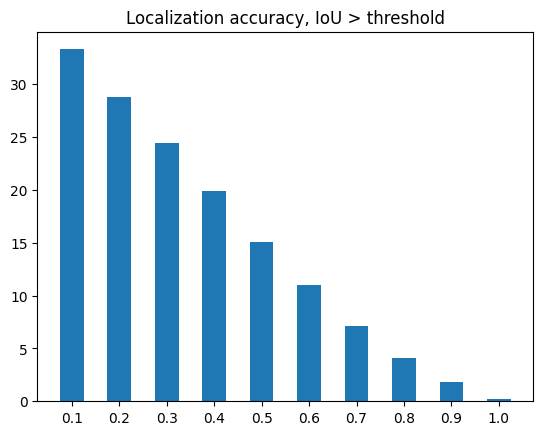}
	\caption{\textbf{Moment retrieval (quantitative):}
            We plot the localization accuracy (y-axis) with various IoU thresholds (x-axis).
	}
	\label{fig:localization_acc}
\end{figure}

\begin{figure*}
	\centering
	\includegraphics[width=0.99\linewidth]{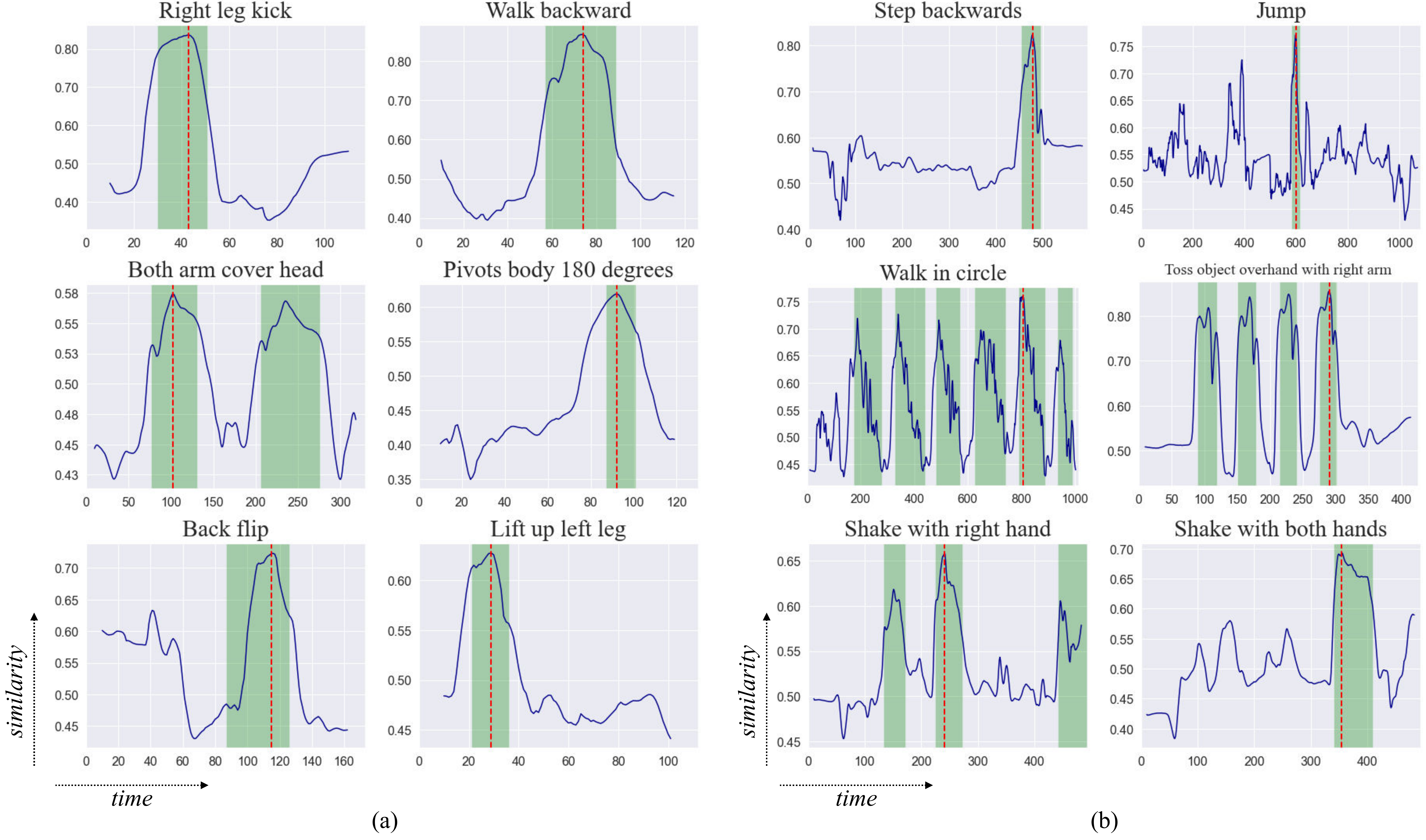} \\
	\caption{\textbf{Moment retrieval (qualitative):}
        To complement
        \if\sepappendix1{Figure 4}
        \else{Figure~\ref{fig:usecases}}
        \fi
        of the main paper,
        (a) we provide six additional temporal localization results
        for various text queries on the BABEL dataset.
        (b) We further visualize six challenging examples
        when querying on very long motion sequences,
        i.e., more than 500 frames (25 seconds).
	}
	\label{fig:localization}
\end{figure*}

For the qualitative results, we display the similarity, centered for each frame, for a window size of 20 frames.
Here, we also implement a temporal pyramid approach,
where we use a sliding window, with window sizes varying between 
10 and 60 frames, and a stride of 5 frames. 
For quantitative evaluation, 
we first obtain the predicted localization 
by selecting the window size and location that 
gives the best similarity with the text query.
Then, we compute the temporal IoU (intersection over union) between the ground-truth 
segment and the predicted one. 
In Figure~\ref{fig:localization_acc}, we report 
the localization accuracy, where a segment is counted
as positive when it has an IoU
more than a given threshold. 
We see that this simple approach can achieve reasonable results (20\% of accuracy, with a threshold of 0.4). \cam{With a fixed window size of [20, 40, 60] frames, we obtain [17\%, 19\%, 14\%], respectively. A dedicated localization method may consider moment proposal generation as in prior video localization work \cite{soldan2021vlg,Soldan_2022_CVPR}, or a proposal-free approach that trains directly to regress temporal boundaries.
}

\section{Additional qualitative results}
\label{appendix:qualitative}

\begin{figure*}
	\centering
\includegraphics[width=0.95\linewidth]{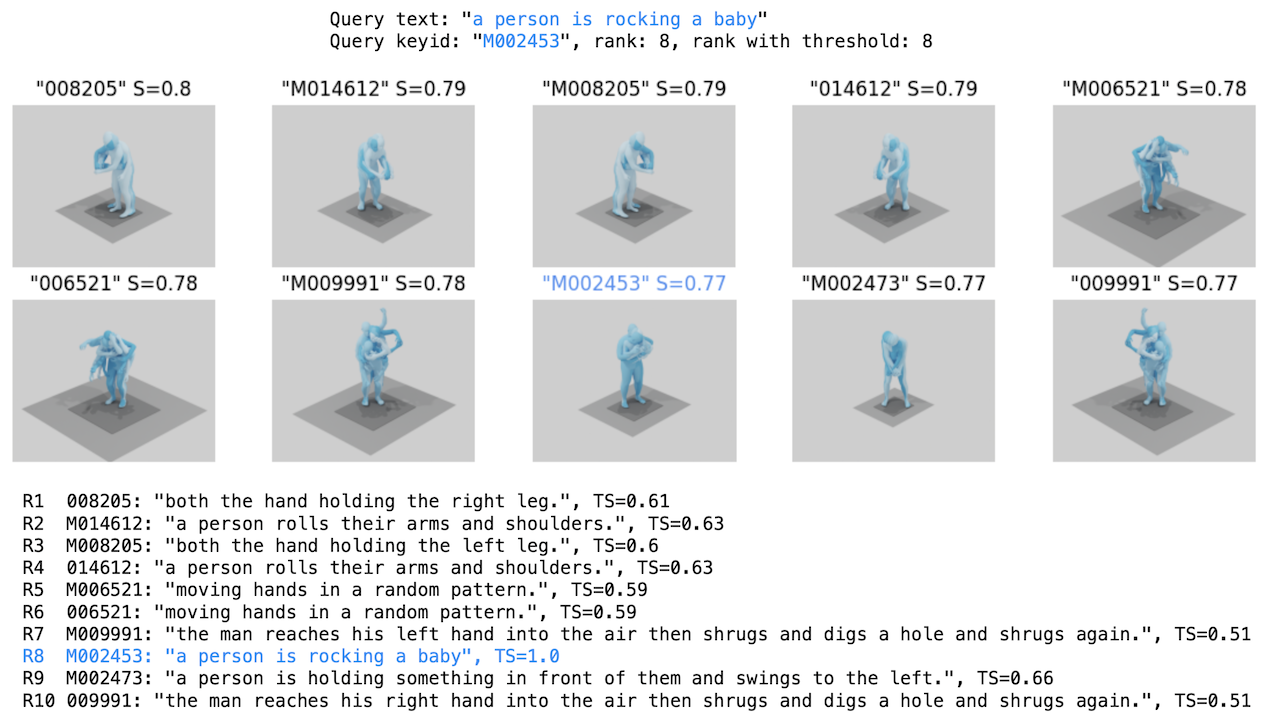}\vspace{0.15cm}
\includegraphics[width=0.95\linewidth]{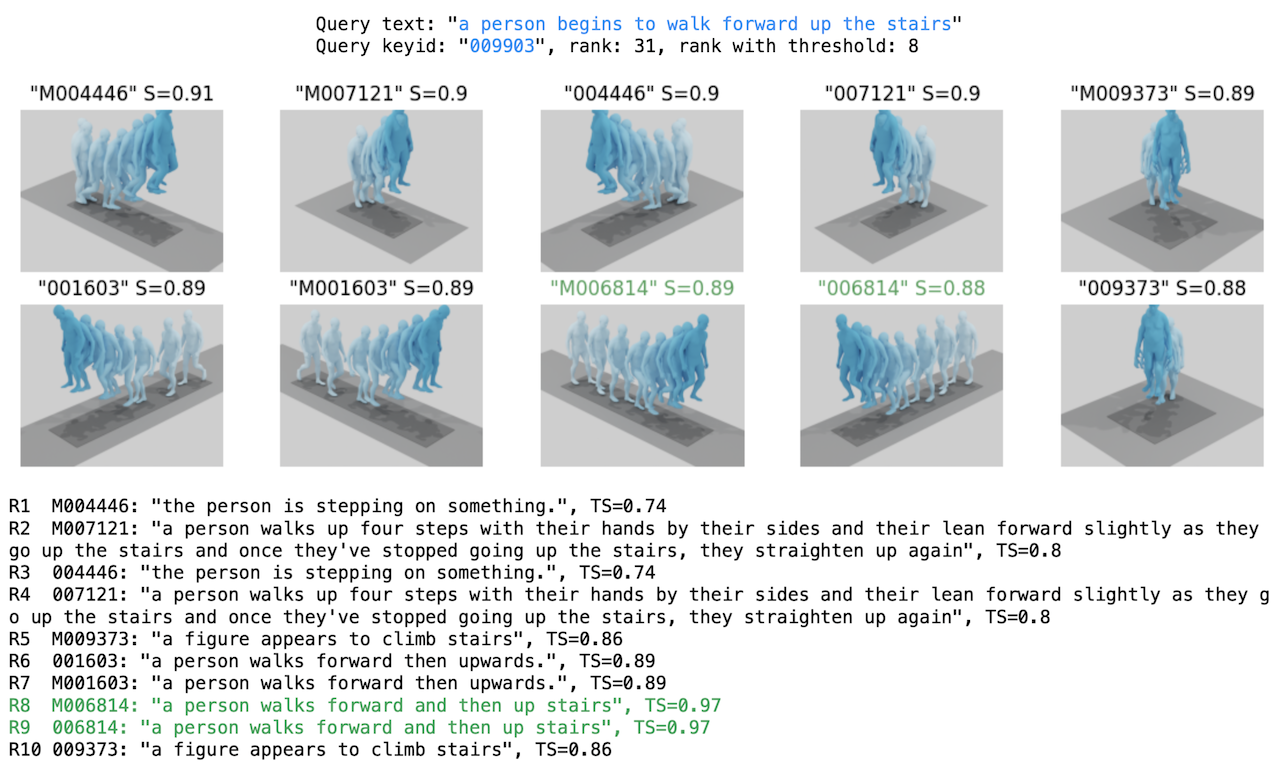}

\caption{\textbf{Protocols (a) and (b) using all 4,380 motions in H3D:}
For each text query, we show the top 10 ranks for the text-to-motion retrieval.
Our model generalizes to the concept of ``rocking a baby"
in the first example, even though this exact same text was not seen in the training set.
In the second example, our model retrieves motions that are all coherent with the input query. However, according to evaluation protocol (a), the correct motion is ranked at 31. With the permissive protocol (b), we mark the rank 8 as correct, because their text similarity (TS) is higher than the threshold 0.95.
} 
	\label{fig:supmat:qual:Ha1}
\end{figure*}

\begin{figure*}
	\centering
\includegraphics[width=0.95\linewidth]{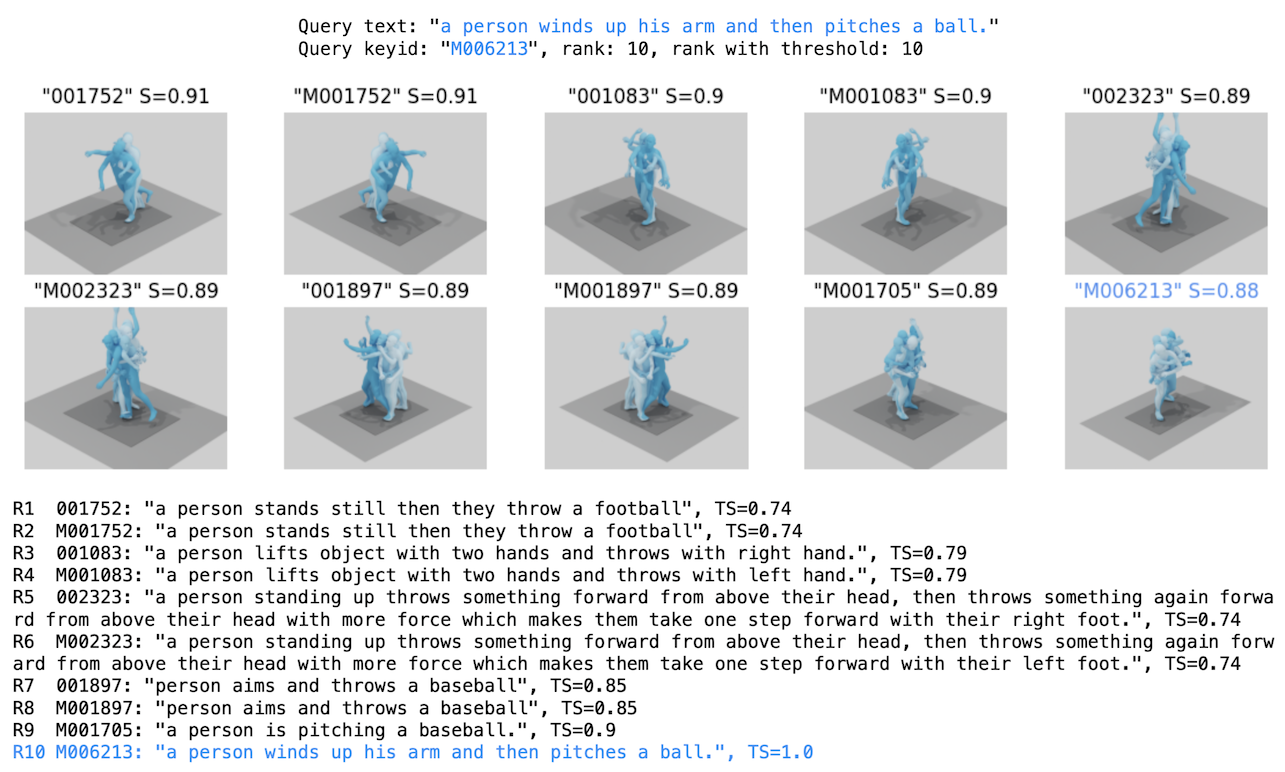}\vspace{0.15cm}

\includegraphics[width=0.95\linewidth]{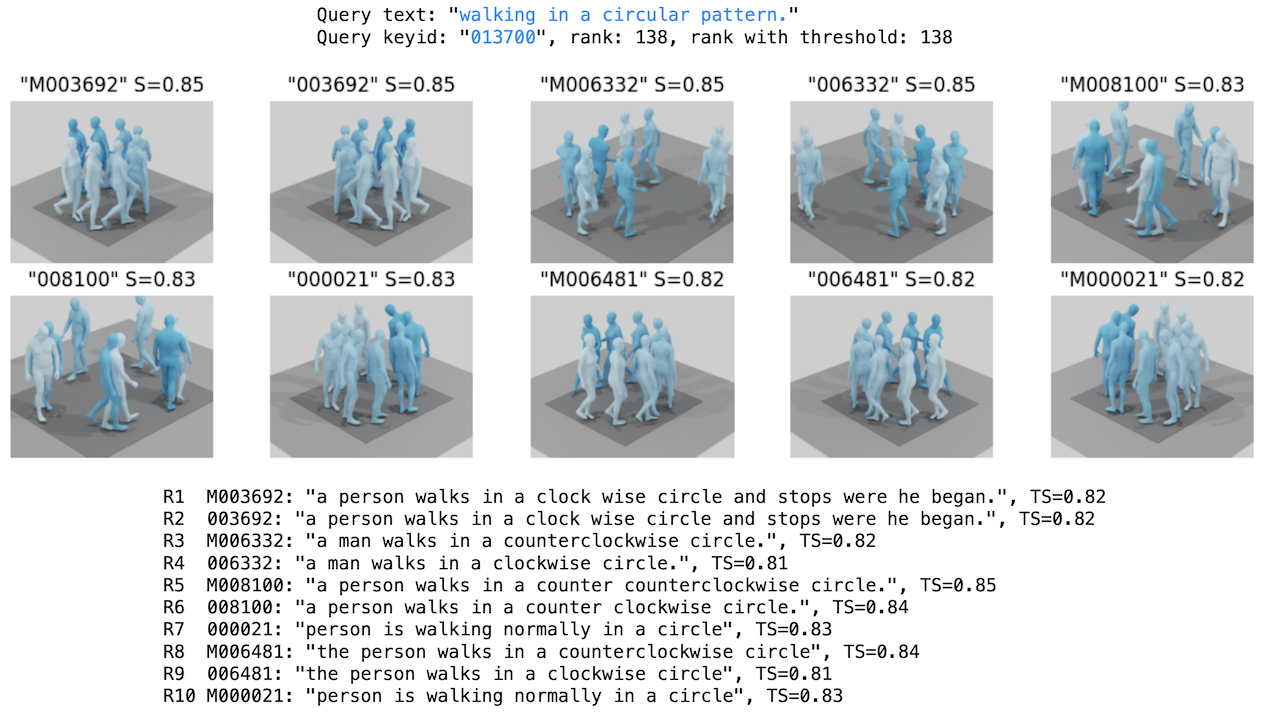}
	\caption{\textbf{Protocols (a) and (b) using all 4,380 motions in H3D (continued):} 
 On both examples, we see that our model retrieves reasonable motions, although the correct motions are ranked at 10 and 138.
 }
	\label{fig:supmat:qual:Ha2}
\end{figure*}

\begin{figure*}
	\centering
\includegraphics[width=0.95\linewidth]{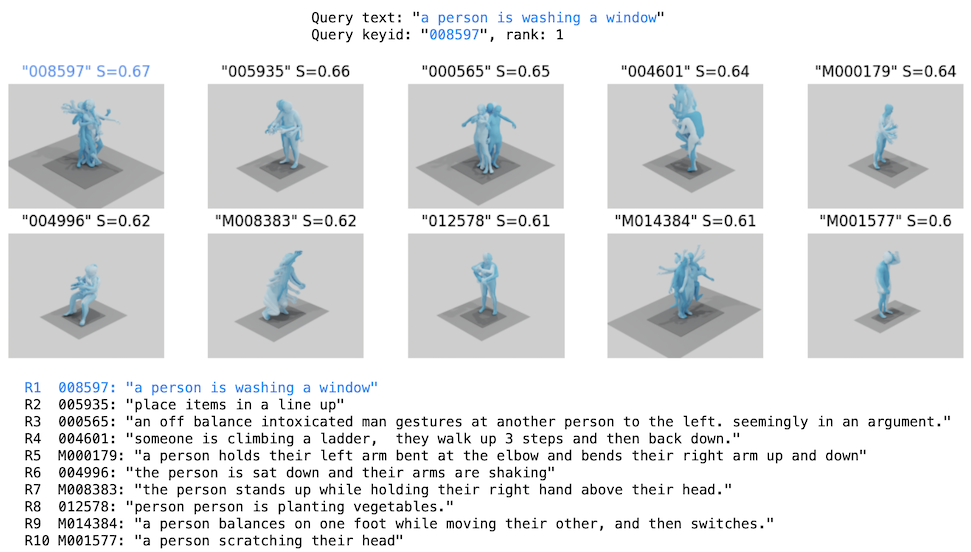}\vspace{0.15cm}
\includegraphics[width=0.95\linewidth]{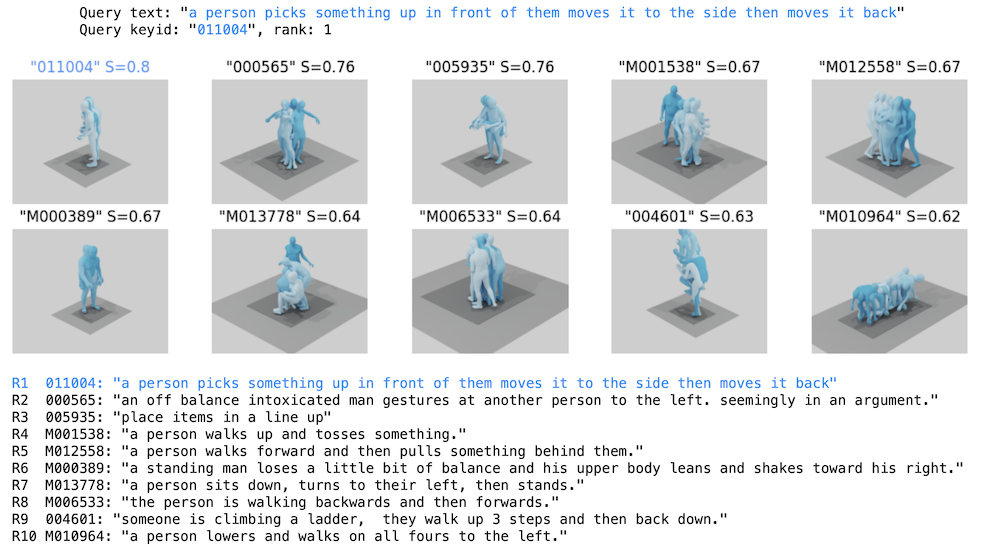}
	\caption{\textbf{Protocol (c) using the most dissimilar 100 texts on H3D:}     
 As there are fewer motions than in protocols (a)(b), and they are more likely to be different, we naturally observe a better performance.
 }
	\label{fig:supmat:qual:Hc}
\end{figure*}

\begin{figure*}
	\centering
\includegraphics[width=0.95\linewidth]{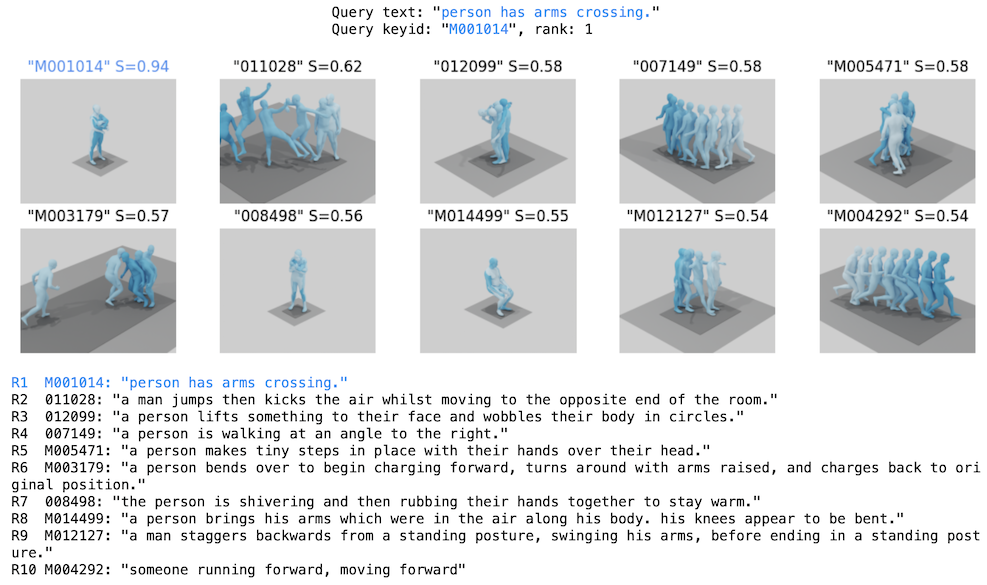}\vspace{0.15cm}
\includegraphics[width=0.95\linewidth]{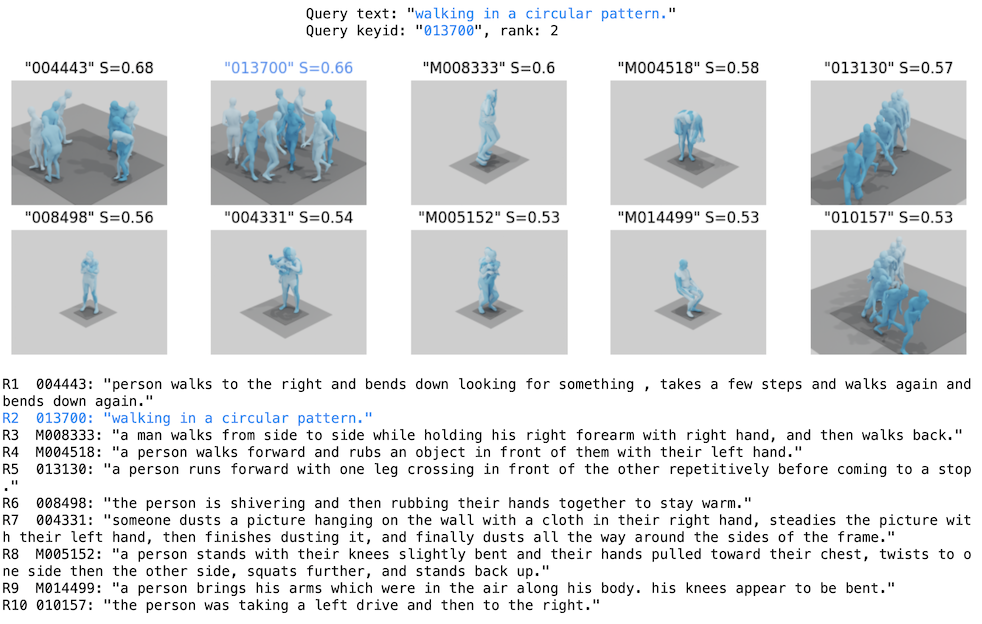}
	\caption{\textbf{Protocol (d) using random batches of size 32 on H3D:}
    As the gallery is very small, the correct
    motion tends to be at top ranks.
 }
	\label{fig:supmat:qual:Hd}
\end{figure*}

In this section, we show qualitative results 
on the challenging H3D dataset
for text-to-motion retrieval on the 4 proposed 
protocols described in
\if\sepappendix1{Section 4.1}
\else{Section~\ref{subsec:datasets}}
\fi
of the main paper.
Protocols (a)(b) are used
in Figures~\ref{fig:supmat:qual:Ha1} and \ref{fig:supmat:qual:Ha2};
(c) in Figure~\ref{fig:supmat:qual:Hc};
and (d) in Figure~\ref{fig:supmat:qual:Hd}.
To reiterate, protocols (a) and (b) use
all the test set (4380 motions) as gallery,
but (b) marks a rank correct if the text similarity
is above a threshold of 0.95.
Protocol (c) considers the  most dissimilar text subset 
of 100 motions.
Protocol (d) is reported for completeness;
it follows \cite{Guo_2022_CVPR}, and 
randomly samples batches of 32 motions.
All examples are randomly chosen,
(i.e., not cherry picked);
therefore, are representative of the
corresponding protocols.

Overall, we observe that our model is
capable of retrieving motions
that are semantically similar to the
text descriptions. The performance
naturally improves as we move from
harder to easier protocols. %
Our detailed observations can be found
in the respective figure captions.

\end{document}